\documentclass[11pt]{article}

\usepackage[final]{acl}

\usepackage{times}
\usepackage{latexsym}
\usepackage[T1]{fontenc}
\usepackage[utf8]{inputenc}

\usepackage{microtype}

\usepackage{amsmath}
\usepackage{amssymb}
\usepackage{amsfonts}

\usepackage{inconsolata}

\usepackage{graphicx}
\usepackage{booktabs}
\usepackage{multirow}
\usepackage{tabularx}
\usepackage[table]{xcolor}

\usepackage{siunitx}

\usepackage{enumitem}

\usepackage{listings}

\usepackage[most]{tcolorbox}
\definecolor{mydarkgreen}{RGB}{0,100,0}
\tcbuselibrary{breakable}

\usepackage{makecell}

\usepackage{pifont}
\usepackage{subcaption}

\definecolor{oursbg}{RGB}{245, 249, 255}
\definecolor{oursbg1}{RGB}{252, 247, 240}
\definecolor{oursbg2}{RGB}{238, 247, 240}
\definecolor{headergray}{RGB}{245,245,245}
\definecolor{inc}{RGB}{0,150,0}
\definecolor{dec}{RGB}{180,0,0}
\definecolor{vanillacol}{RGB}{120,120,120}
\definecolor{ourstext}{RGB}{0,92,175}

\usepackage{pifont}

\newcommand{\internship}{\textsuperscript{*}}
\newcommand{\projectlead}{$^\dagger$}
\newcommand{\corresponding}{$^\ddagger$}

\title{Breaking Block Boundaries: Anchor-based History-stable Decoding for Diffusion Large Language Models}

\author{
 \textbf{Shun Zou\textsuperscript{1,2}\internship},
 \textbf{Yong Wang\textsuperscript{2}\projectlead\corresponding},
 \textbf{Zehui Chen\textsuperscript{1}},
 \textbf{Lin Chen\textsuperscript{1}},\\
 \textbf{Chongyang Tao},
 \textbf{Feng Zhao\textsuperscript{1}\corresponding},
 \textbf{Xiangxiang Chu\textsuperscript{2}}
\\
 \textsuperscript{1}MoE Key Lab of BIPC, University of Science and Technology of China\\
 \textsuperscript{2}AMAP, Alibaba Group
 \\
 \href{https://github.com/zs1314/AHD}{\texttt{https://github.com/zs1314/AHD}}
}

\begin{document}
\maketitle

\renewcommand{\thefootnote}{}
\footnotetext{$^*$ Work done during the internship at AMAP, Alibaba Group. $^\dagger$ Project lead. $^\ddagger$ Corresponding author.}
\renewcommand{\thefootnote}{\arabic{footnote}}

\begin{abstract}
Diffusion Large Language Models (dLLMs) have recently become a promising alternative to autoregressive large language models (ARMs). Semi-autoregressive (Semi-AR) decoding is widely employed in base dLLMs and advanced decoding strategies due to its superior performance. However, our observations reveal that Semi-AR decoding suffers from inherent block constraints, which cause the decoding of many cross-block stable tokens to be unnecessarily delayed. 
To address this challenge, we systematically investigate the identification of stable tokens and present three key findings: (1) naive lookahead decoding is unreliable, (2) token stability closely correlates with convergence trend, and (3) historical information is isolated. Building on these insights, we propose \textbf{A}nchor-based \textbf{H}istory-stable \textbf{D}ecoding (\textbf{AHD}), a training-free, plug-and-play dynamic decoding strategy. Specifically, AHD monitors the stability trend of tokens in real time through dynamic anchors. Once a token reaches stability, it initiates early cross-block decoding to enhance efficiency and performance. Extensive experiments across language, vision-language, and audio-language domains demonstrate that AHD simultaneously improves both performance and inference efficiency. Notably, AHD effectively reverses the performance degradation typically observed in existing advanced decoding acceleration strategies. For instance, on the BBH benchmark, our approach reduces decoding steps by \textbf{80\%} while improving performance by \textbf{3.67\%}.

\end{abstract}

\section{Introduction}
\label{introduction}
Recently, diffusion large language models (dLLMs) have attracted extensive attention from the community and become a strong alternative to autoregressive large language models (ARMs)~\cite{nie2025large,ye2025dream}. In contrast to the next‑token prediction mechanism employed by ARMs, dLLMs utilize bidirectional attention and iterative decoding, facilitating efficient parallel decoding, flexible generation order, and controllable output~\cite{berglund2023reversal}. In recent work, semi‑autoregressive (Semi‑AR) decoding has been widely adopted in base dLLMs~\cite{yang2025mmada,nie2025large,zhu2025llada1.5,bie2025llada2}, and numerous advanced decoding strategies have been developed based on this framework~\cite{wei2025accelerating,wang2025creditdecoding,dong2025saber,wu2025fast,ma2025dinfer,wang2025time}.

\begin{figure}[t]
  \includegraphics[width=\columnwidth]{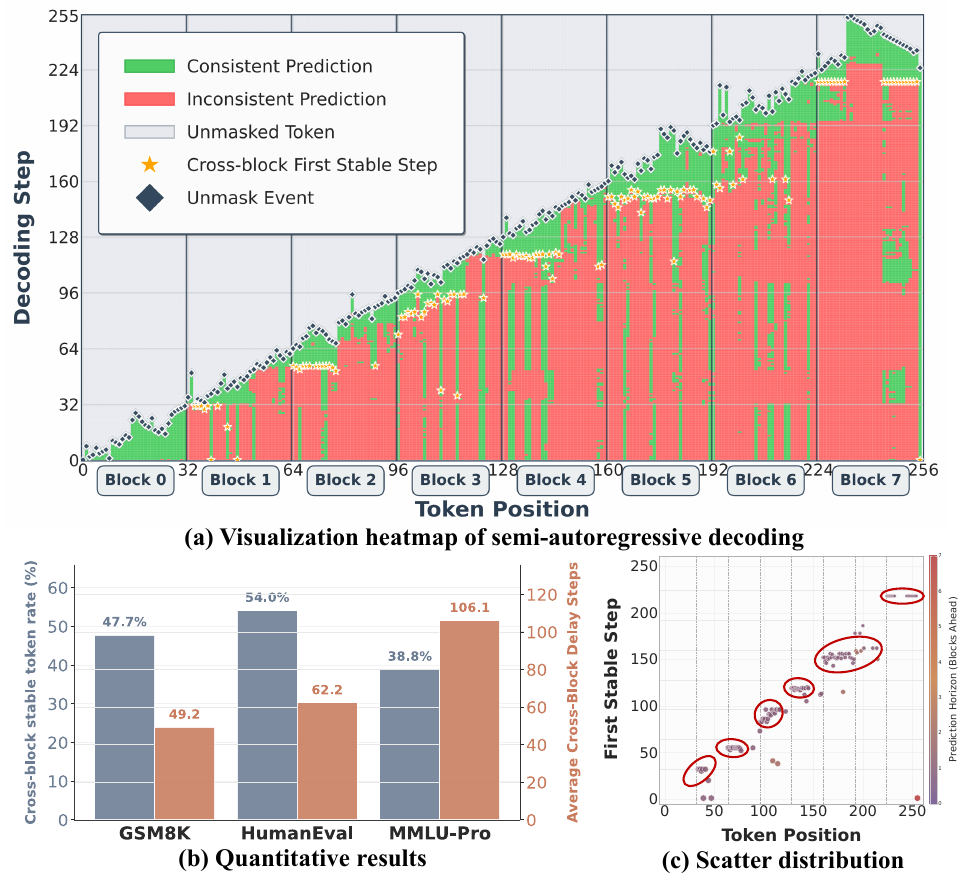}
  \captionsetup{skip=1pt}
  \caption {Systematic analysis of Semi-AR decoding on dLLMs. (a) Heatmap visualization on a sample. Stars indicate the earliest step at which cross-block tokens reach stability, while black squares denote the finally unmasked step. For instance, position 65 is expected to be decoded no earlier than step 64; however, it stabilizes as early as step 53. (b) Qualitative analysis of cross‑block delayed decoding. (c) Distribution of the first stable step.}
    \vspace{-5mm}
  \label{observation1}
\end{figure}

However, Semi‑AR decoding requires the completion of the current block prior to initiating the subsequent one, thereby limiting the potential for parallel decoding. To investigate this constraint, we conducted a systematic study and obtained the following key observations: \textbf{\emph{(i) Block‑Boundary Delay is Prevalent.}} Specifically, Figure~\ref{observation1} (a) and (b) reveal that many tokens already match the final output before their corresponding blocks are decoded and maintain stability until they are eventually unmasked. We define such tokens as cross-block stable tokens. However, the sequential nature of block decoding introduces unnecessary delays for these stable tokens, wasting many decoding steps.
\textbf{\emph{(ii) Cross‑block Stable Tokens Exhibit Radiative Effects.}} Figure~\ref{observation1} (c) illustrates that the initial stable state of a token often emerges in a clustered pattern (i.e., the first stable step).
This indicates that once a token reaches stability, it exerts an influence on neighboring tokens, accelerating their transition into stable states. Conversely, delaying the decoding of a stable token suppresses the entire local region, leading to performance degradation~\cite{lu2025adablock,wang2025time}.


Therefore, we urgently need to free these stable tokens from the sequential block constraint, which is crucial for both inference efficiency and performance. This raises a key question: \textbf{\emph{how can we accurately identify cross‑block stable tokens?}}
Previous studies have often relied on metrics such as confidence and entropy to determine whether a token should be decoded~\cite{nie2025large,wu2025fast,ben2025accelerated}. Therefore, we systematically examine the evolution of the confidence metric for cross‑block stable tokens during decoding, and identify two critical insights: \textbf{\emph{(I) Naive Lookahead Decoding is unreliable.}} As shown in Figure~\ref{observation2} (a), even these tokens that have already reached stability may still experience local fluctuations in subsequent decoding steps. These temporary variations can potentially result in erroneous judgments. \textbf{\emph{(II) Token stability closely correlates with convergence trend.}} As shown in Figure~\ref{observation2} (b), after the first stable step, the confidence trend of these stable tokens changes sharply, exhibiting a rapid growth pattern that we define as the \textbf{\emph{absolute stability trend}}. This motivates us to redefine token stability in terms of the trajectory across decoding steps. Furthermore, it is more critical to capture the emergence of the absolute stability trend at an early stage than solely monitoring peak values. If tokens are unlocked only when their confidence exceeds a high threshold, the decoding process is forced to wait for many unnecessary steps and could exhibit the phenomenon outlined in \textbf{\emph{insight I}}.

Based on the above insights, we identify two essential conditions for the precise detection of cross-block token stability: (1) the formulation of a global stability metric that transcends local behavior; and (2) the capability to dynamically monitor trajectory information. To meet these conditions effectively, the integration of historical information is indispensable.
However, by examining the decoding characteristics of dLLMs, we find that:
\textbf{\emph{(III) Historical information is isolated in standard decoding.}} In conventional iterative decoding within dLLMs, each prediction only depends on the outcome of the immediately preceding step. The final output is determined exclusively by the last decoding step, with no consideration of earlier iterations. Therefore, we argue that breaking this historical isolation by integrating trajectory information is essential for identifying these stable tokens.

In response to these insights, we propose \textbf{A}nchor‑based \textbf{H}istory‑stable \textbf{D}ecoding (\textbf{AHD}), a training‑free, plug‑and‑play dynamic decoding approach for monitoring cross‑block stability.
Specifically, at each decoding step, we take the current step as a dynamic anchor and retrospectively examine prior steps in the historical buffer. Through this process, we establish a stability criterion grounded in historical consistency.
This allows us to capture the absolute stability trend at an early stage, avoiding additional decoding steps, and improving robustness to local fluctuations. Moreover, the early unlocking of these cross-block stable tokens preserves the coherence of the decoding, thereby enhancing generation quality.
Our contributions are summarized as follows: 
\begin{itemize}
[leftmargin=*,topsep=1pt,itemsep=0.4pt]
\item We systematically analyze the key limitations of Semi‑AR decoding and derive three critical insights, which serves as the motivation and key evidence for determining cross‑block stability.
\item Based on the analysis, we propose AHD, a novel cross‑block stability decoding method. It uses dynamic anchors to trace back historical trajectories and capture absolute stability trends, enabling early unlocking of cross-block stable tokens. We also provide theoretical proof of the method.
\item Extensive experiments on language, vision-language, and audio-language domins show that AHD achieves consistent improvements in both performance and inference efficiency, significantly surpassing previous decoding strategies.
\end{itemize}

\section{Preliminary}
\label{preliminary}
In this section, we present an overview of the inference process in dLLMs. Due to space constraints, additional related work is provided in Appendix~\ref{related work}.

\paragraph{Inference Process of dLLMs.}
During inference, diffusion large language models (dLLMs) generate a sequence by iteratively denoising from an initial all‑mask state~\cite{nie2025large}. Given a target sequence length \(L\), the model initializes
\begin{equation}
    \mathbf{x}^T = (\texttt{[MASK]},\dots,\texttt{[MASK]}) \in \mathcal{V}^L,
\end{equation}
and performs \(T\) reverse diffusion steps to reconstruct the final output \(\mathbf{x}^0\).
At each timestep \(t\), the model performs a forward pass on the current state \(\mathbf{x}^t\), yielding token distributions \(\mathbf{p}^t_j\) over the vocabulary for each position \(j = 1,\dots,L\). Subsequently, a confidence‑aware remasking mechanism selects a subset of positions \(\mathcal{M}_{t-1}\) such that:
\begin{equation}
    |\mathcal{M}_{t-1}| \approx \alpha_{t-1} L,
\end{equation}
where \(\alpha_{t-1}\in[0,1]\) is the target masking ratio at step \(t-1\). For each position \(j\), a confidence score is computed by $ s^t_j = \max_{v \in \mathcal{V}} \mathbf{p}^t_j(v)$.

Positions with the lowest confidence scores are iteratively added to the masked token set \(\mathcal{M}_{t-1}\) until the desired number of masks is reached. The remaining positions are updated to their predicted tokens (e.g., via greedy decoding). This process is repeated for each step until \(t = 0\), at which point the final sequence \(\mathbf{x}^0\) is obtained.

\noindent\textbf{Semi-autoregressive Decoding.}
To further enhance quality and controllability, the Semi‑AR decoding has been widely adopted in base dLLMs~\cite{nie2025large,yang2025mmada,zhu2025llada1.5,wang2025diffusion,xin2025lumina}. Recent advances in decoding algorithms often build upon this paradigm~\cite{wu2025fast,dong2025saber,wang2025creditdecoding,wei2025accelerating,kong2025accelerating,kim2025klass,li2025diffusion,ma2025dinfer}. Its core principle is to partition the output sequence into multiple blocks and decode them sequentially from left to right. Within each block, the aforementioned diffusion decoding strategy is applied. Although this method constrains the generation order, it consistently achieves superior performance across various tasks.


\section{Method}
\subsection{Key Limitation of Semi-AR Decoding}
Semi-AR decoding has been widely adopted across various base dLLMs and advanced decoding strategies. However, its architecture supports parallel generation only within individual blocks, while requiring sequential processing across blocks. This structural constraint inherently limits the parallel decoding potential of dLLMs. To address this, we first conducted a systematic analysis of Semi‑AR decoding and identified several key observations, which serve as the motivation and foundation of our proposed method.

\paragraph{Core observation 1: Block-boundary Delay is Prevalent.}
First, we conducted experiments using LLaDA‑8B‑Instruct~\cite{nie2025large} on a representative sample from GSM8K~\cite{cobbe2021training}. As shown in Figure~\ref{observation1} (a), we observe that many tokens already match their final predicted values before their corresponding blocks are decoded, and these tokens remain unchanged until eventually being unmasked. We refer to such tokens as \emph{cross-block stable tokens}.  
To further quantify this phenomenon, we further evaluated three widely-used benchmarks~\cite{chen2021codex,wang2024mmlu,cobbe2021training}. 
As shown in Figure~\ref{observation1} (b), all three benchmarks contain many tokens whose decoding is delayed by block boundaries, resulting in significant waste of decoding steps.


\paragraph{Core observation 2: Cross-block Stable Tokens Exhibit Radiative Effects.}
As shown in Figure~\ref{observation1} (c), we further visualize the temporal and spatial distribution of the \emph{first stable step} (marked as stars in Figure~\ref{observation1} (a)). It is defined as the earliest decoding step at which a token reaches stability. We observe that the onset of stable tokens is not randomly dispersed but instead forms noticeable clusters. Specifically, once a token achieves stability, adjacent tokens tend to stabilize in subsequent steps.
We interpret these findings as follows: during iterative decoding, stable tokens function as semantic anchors, exerting radiative and attractive influences on surrounding tokens. The delayed decoding of these stable tokens suggests that block boundaries disrupt local inference within the block.



\subsection{Prerequisites for Cross-Block Decoding}
Based on above analysis, it is evident that Semi‑AR decoding incurs significant block‑boundary delays.  To address this, a natural objective is to accelerate the release of stable tokens, which are delayed during decoding, thereby reducing overall decoding steps and improving inference coherence. This raises a key question: \textbf{\emph{How can cross-block stable tokens be effectively and accurately identified?}}


\begin{figure}[t]
  \includegraphics[width=\columnwidth]{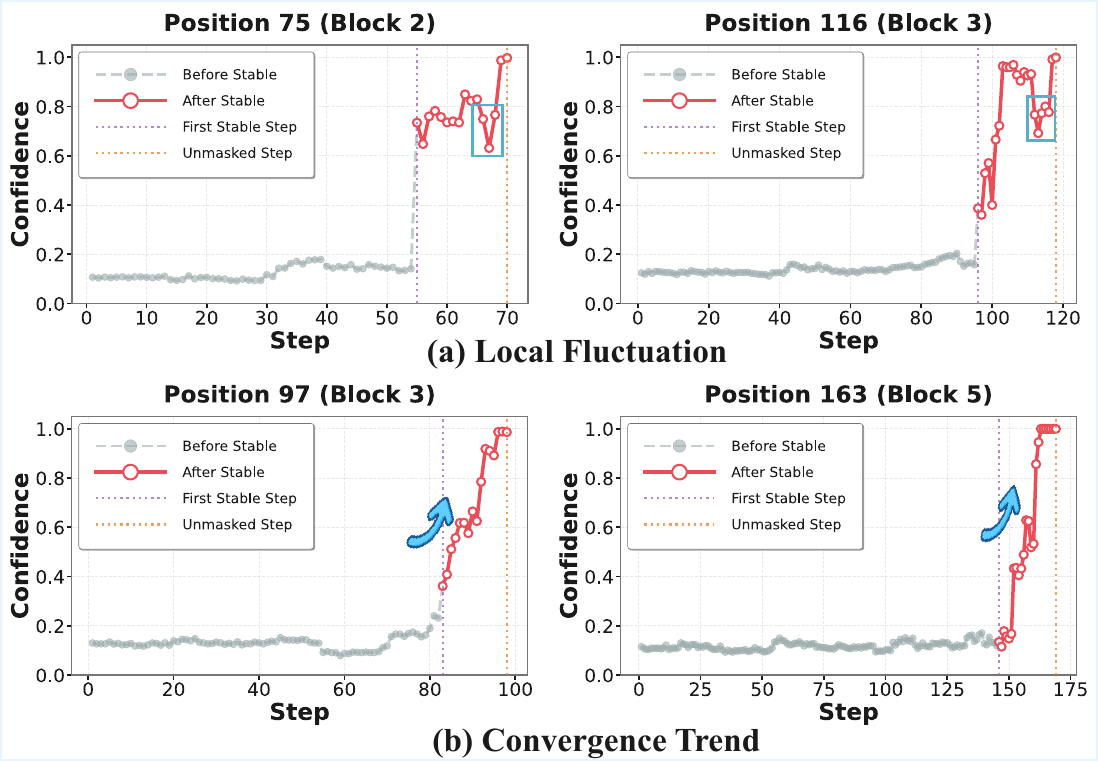} 
  \captionsetup{skip=1pt}
  \caption {Dynamics of confidence during the decoding. (a) Local fluctuations persist even after the token has become stable. (b) Prior to stabilization, the confidence remains flat or oscillatory; however, after crossing the first stable step, it exhibits a marked upward trend.}
  \label{observation2}
  \vspace{-4mm}
\end{figure}

\paragraph{Core insight 1: Naive Lookahead Decoding is  Unreliable.} 
Previous studies have frequently employed confidence and entropy as decoding indicators~\cite{wu2025fast,ben2025accelerated}. Here, we utilize the confidence metric as stability measures for cross-block unlocking, and track the evolution of it. As shown in Figure~\ref{observation2} (a), our observation reveals that stable tokens can still experience brief, large fluctuations in confidence metric even after their first stable step. These low-frequency fluctuations are critical, as they can induce misjudgments, and propagate erroneous signals to subsequent iterations, ultimately affecting the quality of the final output. Moreover, both confidence and entropy metrics are computed based on a single decoding step, rendering them particularly sensitive to local fluctuations.

\vspace{-1mm}
\paragraph{Core insight 2: Token Stability Closely Correlates
with Convergence Trend.}
In addition to local fluctuations, we also identify another common and critical phenomenon. As illustrated in Figure~\ref{observation2} (b), the confidence curves for cross-block stable tokens typically remain persistently low or oscillatory prior to the first stable step, indicating that the position is still engaged in iterative inference. 
However, after exceeding this step, the curves enter a distinct “\emph{convergence phase}”, characterized by a sharp and monotonic increase in confidence. 
We term this phenomenon the \emph{\textbf{absolute stability trend}}. If a token remains unlocked until its confidence exceeds an extremely high threshold, this leads to a significant increase in decoding steps and could exhibit the phenomenon outlined in \textbf{insight 1}. To address this, we redefine token stability based on its confidence trajectory throughout the decoding process. Specifically, our approach is to identify and capture the trajectory at the earliest stage when it exhibits an absolutely stable trend.



\begin{figure*}[t]
  \includegraphics[width=1\textwidth]{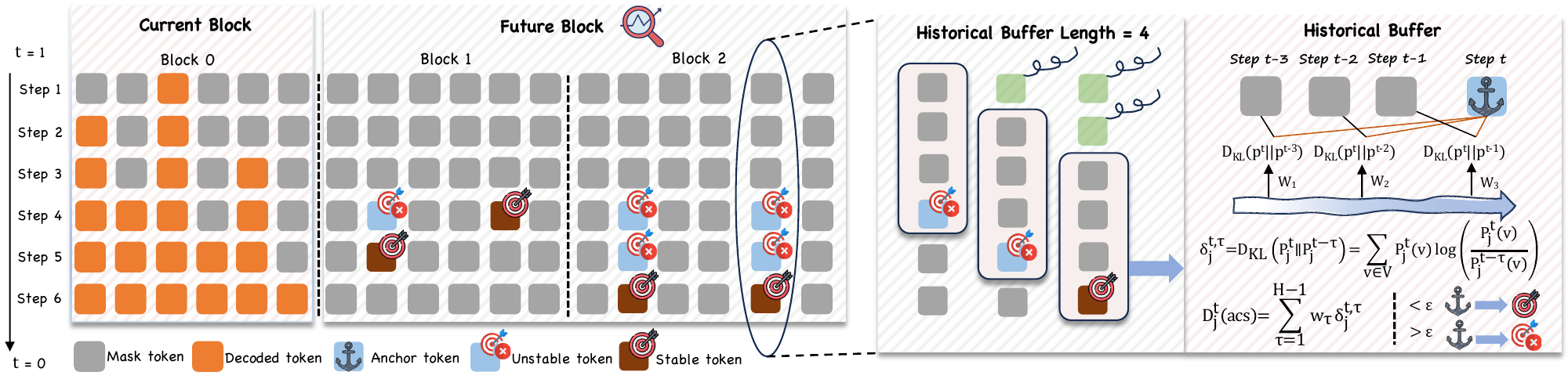} 
  \captionsetup{skip=1pt}
  \caption {Illustration of Anchor‑based History‑stable Decoding (AHD). AHD retrospectively tracks historical trajectories from anchor points to dynamically monitor the absolute stability trend. Cross‑block stable tokens are unlocked early in the absolute stability trend, reducing decoding steps while further unleashing the potential of parallel diffusion decoding and achieving significant performance gains.}
  \label{main_fig}
  \vspace{-5mm}
\end{figure*}

\paragraph{Core insight 3: Historical Information is Isolated in Standard Decoding.}
Building upon above insights, we conclude that two essential conditions must be satisfied to effectively and efficiently detect token stability: (1) the adoption of a global stability metric that transcends single-step evaluation, and (2) the ability to dynamically monitor information and capture the absolute stability trend. 
To satisfy these conditions, a crucial consideration is the incorporation of historical information. Through analysis of the standard decoding in dLLMs, we observe that each inference step operates independently, disregarding previous predictions. Consequently, the final output is completely determined by the prediction \( p_\theta(x^0 \mid x^1) \) at the last step, while all intermediate predictive distributions during decoding are ignored. Therefore, it is essential to break this inherent characteristic and leverage historical trajectory information.  




\subsection{Anchor-based History-stable Decoding}
\label{AHD}
Building on the above analysis, we propose \textbf{A}nchor-based \textbf{H}istory-stable \textbf{D}ecoding (\textbf{AHD}), a training-free, plug-and-play decoding strategy. 
The core idea is to use dynamic anchors to compute a historical consistency stability criterion and capture the \emph{\textbf{absolute stability trend}} at an early stage. This safely unlocks the potential of cross-block parallel decoding and improves the inference coherence.


As shown in Figure~\ref{main_fig}, at each decoding step \( t \), we first divide the sequence \( X^{t} \) into the current block \( B^{t}_{\text{current}} \) and the future block \( B^{t}_{\text{future}} \). In traditional Semi-AR decoding, token generation at step $t$ is restricted to \( B^{t}_{\text{current}} \). Our approach adopts the confidence-aware parallel decoding strategy to define a subset \( G^{t}_{c} \) within \( B^{t}_{\text{current}} \). Simultaneously, within \( B^{t}_{\text{future}} \), we employ same confidence-aware parallel decoding strategy and AHD to construct a cross-block lookahead decoding subset \( G^{t}_{f} \). 
To capture the dynamic evolution of token distributions across decoding steps, we utilize KL divergence as a metric for distributional change. Specifically, for position \(j \) at decoding step \(t\), we define the one-step KL divergence \( D^t_{j}(\mathrm{one\mbox{-}step}) \), as follows:
\begin{equation}
\begin{aligned}
    D^t_{j}(\mathrm{one\mbox{-}step}) & \triangleq D_{KL}(P_{j}^{t}||P_{j}^{t-1}) \\
     & =\sum_{v\in V}P_{j}^{t}(v)\log\frac{P_{j}^{t}(v)}{P_{j}^{t-1}(v)}
\end{aligned}
\end{equation}
Here, \( P^{t}_{j}(v) \) and \( P^{t-1}_{j}(v) \) represent the probabilities of generating token \( v \in V \) at position \( j \) for decoding steps \( t \) and \( t-1 \), where  \( V \) is the vocabulary.
However, as demonstrated by our \textbf{core insight 1}, \(D(\mathrm{one\mbox{-}step}) \) is highly sensitive to local fluctuations, which can lead to misjudgments.


\begin{table*}[t]
    \centering
    \resizebox{1\textwidth}{!}{\begin{tabular}{llcccccc}
        \toprule
        \rowcolor{headergray}
        \textbf{Task} & \textbf{Metric} 
        & \textbf{Vanilla} 
        & \textbf{PC-sampler} 
        & \textbf{Fast-dLLM} 
        & \textbf{KLASS} 
        & \textbf{Saber} 
        & \cellcolor{oursbg}\textbf{Ours} \\
        \midrule
        \multirow{2}{*}{\textbf{HumanEval}} 
        & \emph{Score} ↑  & {\color{vanillacol}40.85} 
                  & 40.24 {\scriptsize\textcolor{dec}{(-0.61)}} 
                  & 41.46 {\scriptsize\textcolor{inc}{(+0.61)}} 
                  & 40.85 {\scriptsize\textcolor{dec}{(0.00)}} 
                  & 36.59 {\scriptsize\textcolor{dec}{(-4.26)}} 
                  & \cellcolor{oursbg}\textbf{43.29} {\scriptsize\textcolor{inc}{(+2.44)}} \\
        & \emph{Step} ↓   & {\color{vanillacol}256} 
                  & 256 
                  & 78.36 {\scriptsize\textcolor{inc}{($-69\%$)}} 
                  & 131.48 {\scriptsize\textcolor{inc}{($-49\%$)}} 
                  & 82.83 {\scriptsize\textcolor{inc}{($-68\%$)}} 
                  & \cellcolor{oursbg}\textbf{77.24} {\scriptsize\textcolor{inc}{($-70\%$)}} \\
        \midrule
        \multirow{2}{*}{\textbf{MBPP}} 
        & \emph{Score} ↑  & {\color{vanillacol}29.20} 
                  & 31.20 {\scriptsize\textcolor{inc}{(+2.00)}} 
                  & 29.40 {\scriptsize\textcolor{inc}{(+0.20)}} 
                  & 29.20 {\scriptsize\textcolor{dec}{(0.00)}} 
                  & 26.00 {\scriptsize\textcolor{dec}{(-3.20)}} 
                  & \cellcolor{oursbg}\textbf{31.20} {\scriptsize\textcolor{inc}{(+2.00)}} \\
        & \emph{Step} ↓   & {\color{vanillacol}256} 
                  & 256 
                  & 70.05 {\scriptsize\textcolor{inc}{($-73\%$)}} 
                  & 113.26 {\scriptsize\textcolor{inc}{($-56\%$)}} 
                  & 71.79 {\scriptsize\textcolor{inc}{($-72\%$)}} 
                  & \cellcolor{oursbg}\textbf{66.03} {\scriptsize\textcolor{inc}{($-74\%$)}} \\
        \midrule
        \multirow{2}{*}{\textbf{BBH}} 
        & \emph{Score} ↑  & {\color{vanillacol}53.11} 
                  & 48.55 {\scriptsize\textcolor{dec}{(-4.56)}} 
                  & 53.17 {\scriptsize\textcolor{inc}{(+0.06)}} 
                  & 53.03 {\scriptsize\textcolor{dec}{(-0.08)}} 
                  & 52.88 {\scriptsize\textcolor{dec}{(-0.23)}} 
                  & \cellcolor{oursbg}\textbf{56.78} {\scriptsize\textcolor{inc}{(+3.67)}} \\
        & \emph{Step} ↓   & {\color{vanillacol}256} 
                  & 256 
                  & 55.85 {\scriptsize\textcolor{inc}{($-78\%$)}} 
                  & 98.04 {\scriptsize\textcolor{inc}{($-62\%$)}} 
                  & 87.31 {\scriptsize\textcolor{inc}{($-66\%$)}} 
                  & \cellcolor{oursbg}\textbf{51.48} {\scriptsize\textcolor{inc}{($-80\%$)}} \\
        \midrule
        \multirow{2}{*}{\textbf{MMLU-Pro}} 
        & \emph{Score} ↑  & {\color{vanillacol}35.57} 
                  & 33.90 {\scriptsize\textcolor{dec}{(-1.67)}} 
                  & 35.70 {\scriptsize\textcolor{inc}{(+0.13)}} 
                  & 35.67 {\scriptsize\textcolor{inc}{(+0.10)}} 
                  & 36.10 {\scriptsize\textcolor{inc}{(+0.53)}} 
                  & \cellcolor{oursbg}\textbf{37.42} {\scriptsize\textcolor{inc}{(+1.85)}} \\
        & \emph{Step} ↓   & {\color{vanillacol}256} 
                  & 256 
                  & 134.76 {\scriptsize\textcolor{inc}{($-47\%$)}} 
                  & 187.17 {\scriptsize\textcolor{inc}{($-27\%$)}} 
                  & \textbf{123.83} {\scriptsize\textcolor{inc}{($-52\%$)}} 
                  & \cellcolor{oursbg}133.06 {\scriptsize\textcolor{inc}{($-48\%$)}} \\
        \midrule
        \multirow{2}{*}{\textbf{TruthfulQA}} 
        & \emph{Score} ↑  & {\color{vanillacol}40.39} 
                  & 40.39 {\scriptsize\textcolor{dec}{(0.00)}} 
                  & 40.39 {\scriptsize\textcolor{dec}{(0.00)}} 
                  & 40.39 {\scriptsize\textcolor{dec}{(0.00)}} 
                  & 40.76 {\scriptsize\textcolor{inc}{(+0.37)}} 
                  & \cellcolor{oursbg}\textbf{41.49} {\scriptsize\textcolor{inc}{(+1.10)}} \\
        & \emph{Step} ↓   & {\color{vanillacol}256} 
                  & 256 
                  & 67.22 {\scriptsize\textcolor{inc}{($-74\%$)}} 
                  & 85.56 {\scriptsize\textcolor{inc}{($-67\%$)}} 
                  & 53.25 {\scriptsize\textcolor{inc}{($-79\%$)}} 
                  & \cellcolor{oursbg}\textbf{52.91} {\scriptsize\textcolor{inc}{($-79\%$)}} \\
        \midrule
        \multirow{2}{*}{\textbf{Math}} 
        & \emph{Score} ↑  & {\color{vanillacol}33.50} 
                  & \textbf{33.62} {\scriptsize\textcolor{inc}{(+0.12)}} 
                  & 33.26 {\scriptsize\textcolor{dec}{(-0.24)}} 
                  & 33.38 {\scriptsize\textcolor{dec}{(-0.12)}} 
                  & 33.48 {\scriptsize\textcolor{dec}{(-0.02)}} 
                  & \cellcolor{oursbg}\textbf{33.62} {\scriptsize\textcolor{inc}{(+0.12)}} \\
        & \emph{Step} ↓   & {\color{vanillacol}256} 
                  & 256 
                  & 96.49 {\scriptsize\textcolor{inc}{($-62\%$)}} 
                  & 156.72 {\scriptsize\textcolor{inc}{($-39\%$)}} 
                  & 96.41 {\scriptsize\textcolor{inc}{($-62\%$)}} 
                  & \cellcolor{oursbg}\textbf{88.50} {\scriptsize\textcolor{inc}{($-65\%$)}} \\
        \midrule
        \multirow{2}{*}{\textbf{Asdiv}} 
        & \emph{Score} ↑  & {\color{vanillacol}75.57}
                  & 74.19 {\scriptsize\textcolor{dec}{(-1.38)}}
                  & 74.79 {\scriptsize\textcolor{dec}{(-0.78)}}
                  & 75.40 {\scriptsize\textcolor{dec}{(-0.17)}}
                  & 75.18 {\scriptsize\textcolor{dec}{(-0.39)}}
                  & \cellcolor{oursbg}\textbf{77.09} {\scriptsize\textcolor{inc}{(+1.52)}} \\
        & \emph{Step} ↓   & {\color{vanillacol}256}
                  & 256 
                  & 65.57 {\scriptsize\textcolor{inc}{($-74\%$)}} 
                  & 115.35 {\scriptsize\textcolor{inc}{($-55\%$)}} 
                  & 73.37 {\scriptsize\textcolor{inc}{($-71\%$)}} 
                  & \cellcolor{oursbg}\textbf{61.58} {\scriptsize\textcolor{inc}{($-76\%$)}} \\
        \bottomrule
    \end{tabular}
}
\vspace{-2mm}
\caption{Evaluation results of LLaDA-8B-Instruct on seven benchmarks. Values in parentheses denote the performance change relative to the baseline naive decoding and the reduction in decoding steps. The best performance and lowest decoding steps are highlighted in \textbf{bold}.}
\vspace{-5mm}
\label{tab:results-instruct}
\end{table*}


Based on \textbf{observation 2} and \textbf{insights 2 and 3}, we leverage the historical consistency stability criterion to retrospectively monitor the \emph{\textbf{absolute stability trend}} in real time. Specifically, for each position \( j \in B^{t}_{\text{future}} \) within the future block, we maintain a historical buffer of length \( H \), denoted as \(\mathcal{H}^{t}_{j} = \{ P^{t-H+1}_{j}, \dots, P^{t}_{j} \}\). Simultaneously, for each historical buffer, we dynamically designate an anchor point for the current decoding step, denoted as \( P^{t}_{j,anchor} \), which is formally equivalent to \( P^{t}_{j} \). 
The core idea of the strategy is to employ the dynamic anchor as the evaluation baseline, and retrospectively examine all steps within the historical buffer \(\{ P^{t-H+1}_{j}, \dots, P^{t}_{j} \}\), excluding the anchor, to evaluate the absolute stability trend within that historical buffer. We formally define the anchored KL divergence as follows:
\begin{equation}
    \delta _{j}^{t,\tau }\triangleq D_{KL}(P_{j,anchor}^{t}||P_{j}^{t-\tau }).
\end{equation}
Here, \( \tau \in [1, H-1] \) represents the backtracking step size. Accordingly, we obtain a historical consistency sequence \(\{ \delta^{t,1}_{j}, \dots, \delta^{t,H-1}_{j} \}\).  
To aggregate evidence across the entire historical window, we compute a weighted sum of the \(\delta^{t,\tau}_{j}\) within the sequence, resulting in the final \textbf{a}nchored \textbf{c}onsistency \textbf{s}core \( D^{t}_{j}(acs) \).  Formally, this can be expressed as:
\begin{equation}
    D_{j}^{t}(acs)=\sum_{\tau=1}^{H-1}w_{\tau}\delta_{j}^{t,\tau}
\end{equation}
Here, we assign the exponentially decaying weights \( w_{\tau} \) to historical gap \( \tau \), thereby assigning greater importance to more recent history.  Specifically, the weight is defined as:
\begin{equation}
    w_{\tau}=\frac{e^{-\lambda\tau}}{Z},Z=\sum_{k=1}^{H-1}e^{-\lambda k},
\end{equation}
where \( \lambda > 0 \) controls the rate of decay. 
When \( D^{t}_{j}(acs) \) falls below the historical consistency threshold \(\varepsilon\), we consider that, within this historical time window, \(P^{t}_{j,anchor}\) has already reached an absolutely stable trend and will be incorporated into the \(G^{t}_{f}\).
Formally, this subset is defined as:
\begin{equation}
    G_{f}^{t}=\{j|j\in B_{future}\wedge D_{j}^{t}\left(acs\right)<\varepsilon\}
\end{equation}
Finally, we update the sequence \( X^{t} \) at the corresponding decoding positions from \( G^{t}_{\text{unmasked}} \), where \( G^{t}_{\text{unmasked}} \) is the union of \( G^{t}_{c} \) and \( G^{t}_{f} \). 

\section{Experiments}

\subsection{Experimental Setup}
\paragraph{Implementation Details.}
In the language domain, we evaluate the proposed method on two widely used dLLMs: LLaDA-8B-Instruct~\cite{nie2025large} and LLaDA-1.5~\cite{zhu2025llada1.5}, with a generation length of 256 and a block length of 32. In the vision-language and audio-language domains, we use MMaDA-8B-MixCoT~\cite{yang2025mmada} and DIFFA~\cite{zhou2025diffa}, respectively. Additionally, the historical consistency threshold \( \varepsilon \) and the historical buffer length \( \mathcal{H} \) are set to 0.01 and 6 by default. More implementation details are provided in Appendix~\ref{sec:appendix-exper-detils}.



\noindent\textbf{Datasets and Tasks.}
Our evaluation in the language domain encompasses three key capability areas across seven benchmarks. For code generation, we utilize HumanEval~\cite{chen2021codex} and MBPP~\cite{austin2021program}. For general tasks, we employ BBH~\cite{suzgun2023challenging}, MMLU-Pro~\cite{wang2024mmlu}, and TruthfulQA~\cite{lin2022truthfulqa}. For mathematical reasoning, we adopt Math~\cite{hendrycks2021measuring} and Asdiv~\cite{miao2020diverse}. In the vision-language domain, our assessment covers five multimodal benchmarks: MATH-Vision~\cite{wang2024measuring}, MathVista~\cite{lu2023mathvista}, ScienceQA~\cite{lu2022learn}, GQA~\cite{hudson2019gqa} and MME~\cite{fu2023mme}. In the audio-language domain, we conduct evaluations on five tasks from VoiceBench~\cite{chen2024voicebench}. 



\begin{table*}[t]
    \centering
    \resizebox{1\textwidth}{!}{\begin{tabular}{llcccccc}
        \toprule
        \rowcolor{headergray}
        \textbf{Task} & \textbf{Metric} 
        & \textbf{Vanilla} 
        & \textbf{PC-sampler} 
        & \textbf{Fast-dLLM} 
        & \textbf{KLASS} 
        & \textbf{Saber} 
        & \cellcolor{oursbg}\textbf{Ours} \\
        \midrule
        \multirow{2}{*}{\textbf{HumanEval}} 
        & \emph{Score} ↑  & {\color{vanillacol}42.68} 
                  & 31.88 {\scriptsize\textcolor{dec}{(-10.80)}} 
                  & 43.29 {\scriptsize\textcolor{inc}{(+0.61)}} 
                  & 43.29 {\scriptsize\textcolor{inc}{(+0.61)}} 
                  & 42.07 {\scriptsize\textcolor{dec}{(-0.61)}} 
                  & \cellcolor{oursbg}\textbf{43.90} {\scriptsize\textcolor{inc}{(+1.22)}} \\

        & \emph{Step} ↓   & {\color{vanillacol}256} 
                  & 256 
                  & 75.88 {\scriptsize\textcolor{inc}{($-70\%$)}} 
                  & 131.01 {\scriptsize\textcolor{inc}{($-49\%$)}} 
                  & 82.07 {\scriptsize\textcolor{inc}{($-68\%$)}} 
                  & \cellcolor{oursbg}\textbf{75.85} {\scriptsize\textcolor{inc}{($-70\%$)}} \\
        \midrule
        \multirow{2}{*}{\textbf{MBPP}} 
        & \emph{Score} ↑  & {\color{vanillacol}27.20} 
                  & 25.60 {\scriptsize\textcolor{dec}{(-1.60)}} 
                  & 28.00 {\scriptsize\textcolor{inc}{(+0.80)}} 
                  & 27.20 {\scriptsize\textcolor{dec}{(0.00)}} 
                  & 25.40 {\scriptsize\textcolor{dec}{(-1.80)}} 
                  & \cellcolor{oursbg}\textbf{28.60} {\scriptsize\textcolor{inc}{(+1.40)}} \\

        & \emph{Step} ↓   & {\color{vanillacol}256} 
                  & 256 
                  & 73.01 {\scriptsize\textcolor{inc}{($-71\%$)}} 
                  & 119.34 {\scriptsize\textcolor{inc}{($-53\%$)}} 
                  & 77.70 {\scriptsize\textcolor{inc}{($-70\%$)}} 
                  & \cellcolor{oursbg}\textbf{72.88} {\scriptsize\textcolor{inc}{($-72\%$)}} \\
        \midrule
        \multirow{2}{*}{\textbf{BBH}} 
        & \emph{Score} ↑  & {\color{vanillacol}50.35} 
                  & 47.03 {\scriptsize\textcolor{dec}{(-3.32)}} 
                  & 49.85 {\scriptsize\textcolor{dec}{(-0.50)}} 
                  & 50.28 {\scriptsize\textcolor{dec}{(-0.07)}} 
                  & 50.02 {\scriptsize\textcolor{dec}{(-0.33)}} 
                  & \cellcolor{oursbg}\textbf{51.90} {\scriptsize\textcolor{inc}{(+1.55)}} \\

        & \emph{Step} ↓   & {\color{vanillacol}256} 
                  & 256 
                  & 57.89 {\scriptsize\textcolor{inc}{($-77\%$)}} 
                  & 101.11 {\scriptsize\textcolor{inc}{($-61\%$)}} 
                  & 88.90 {\scriptsize\textcolor{inc}{($-65\%$)}} 
                  & \cellcolor{oursbg}\textbf{56.00} {\scriptsize\textcolor{inc}{($-78\%$)}} \\
        \midrule
        \multirow{2}{*}{\textbf{MMLU-Pro}} 
        & \emph{Score} ↑  & {\color{vanillacol}37.15} 
                  & 35.01 {\scriptsize\textcolor{dec}{(-2.14)}} 
                  & 37.21 {\scriptsize\textcolor{inc}{(+0.06)}} 
                  & 37.26 {\scriptsize\textcolor{inc}{(+0.11)}} 
                  & 36.78 {\scriptsize\textcolor{dec}{(-0.37)}} 
                  & \cellcolor{oursbg}\textbf{38.48} {\scriptsize\textcolor{inc}{(+1.33)}} \\

        & \emph{Step} ↓   & {\color{vanillacol}256} 
                  & 256 
                  & 134.72 {\scriptsize\textcolor{inc}{($-47\%$)}} 
                  & 188.14 {\scriptsize\textcolor{inc}{($-27\%$)}} 
                  & \textbf{125.71} {\scriptsize\textcolor{inc}{($-51\%$)}} 
                  & \cellcolor{oursbg}133.02 {\scriptsize\textcolor{inc}{($-48\%$)}} \\
        \midrule
        \multirow{2}{*}{\textbf{TruthfulQA}} 
        & \emph{Score} ↑  & {\color{vanillacol}41.27} 
                  & 41.78 {\scriptsize\textcolor{inc}{(+0.51)}} 
                  & 41.49 {\scriptsize\textcolor{inc}{(+0.22)}} 
                  & 41.37 {\scriptsize\textcolor{inc}{(+0.10)}} 
                  & 40.88 {\scriptsize\textcolor{dec}{(-0.39)}} 
                  & \cellcolor{oursbg}\textbf{41.98} {\scriptsize\textcolor{inc}{(+0.71)}} \\

        & \emph{Step} ↓   & {\color{vanillacol}256} 
                  & 256 
                  & 83.93 {\scriptsize\textcolor{inc}{($-67\%$)}} 
                  & 104.26 {\scriptsize\textcolor{inc}{($-59\%$)}} 
                  & 68.50 {\scriptsize\textcolor{inc}{($-73\%$)}} 
                  & \cellcolor{oursbg}\textbf{72.21} {\scriptsize\textcolor{inc}{($-72\%$)}} \\
        \midrule
        \multirow{2}{*}{\textbf{Math}} 
        & \emph{Score} ↑  & {\color{vanillacol}34.32} 
                  & 34.22 {\scriptsize\textcolor{dec}{(-0.10)}} 
                  & 34.21 {\scriptsize\textcolor{dec}{(-0.11)}} 
                  & 34.30 {\scriptsize\textcolor{dec}{(-0.02)}} 
                  & 34.26 {\scriptsize\textcolor{dec}{(-0.06)}} 
                  & \cellcolor{oursbg}\textbf{34.34} {\scriptsize\textcolor{inc}{(+0.02)}} \\

        & \emph{Step} ↓   & {\color{vanillacol}256} 
                  & 256 
                  & 93.36 {\scriptsize\textcolor{inc}{($-64\%$)}} 
                  & 154.26 {\scriptsize\textcolor{inc}{($-40\%$)}} 
                  & 95.56 {\scriptsize\textcolor{inc}{($-63\%$)}} 
                  & \cellcolor{oursbg}\textbf{92.72} {\scriptsize\textcolor{inc}{($-64\%$)}} \\
        \midrule
        \multirow{2}{*}{\textbf{Asdiv}} 
        & \emph{Score} ↑  & {\color{vanillacol}77.18}
                  & 74.97 {\scriptsize\textcolor{dec}{(-2.21)}}
                  & 76.36 {\scriptsize\textcolor{dec}{(-0.82)}}
                  & 76.79 {\scriptsize\textcolor{dec}{(-0.39)}}
                  & 75.84 {\scriptsize\textcolor{dec}{(-1.34)}}
                  & \cellcolor{oursbg}\textbf{78.79} {\scriptsize\textcolor{inc}{(+1.61)}} \\

        & \emph{Step} ↓   & {\color{vanillacol}256}
                  & 256 
                  & 63.83 {\scriptsize\textcolor{inc}{($-75\%$)}} 
                  & 114.52 {\scriptsize\textcolor{inc}{($-55\%$)}} 
                  & 73.19 {\scriptsize\textcolor{inc}{($-71\%$)}} 
                  & \cellcolor{oursbg}\textbf{60.21} {\scriptsize\textcolor{inc}{($-76\%$)}} \\
        \bottomrule
    \end{tabular}
}
\vspace{-3mm}
\caption{Evaluation results of LLaDA-1.5 on seven benchmarks.}
\label{tab:results-llada1.5}
\vspace{-5mm}
\end{table*}

\begin{table}[ht]
    \centering

    \resizebox{\columnwidth}{!}{
    \begin{tabular}{l l c c}
        \toprule
        \textbf{Task} & \textbf{Method} &
        \emph{\textbf{Score}} ↑ &
        \emph{\textbf{Speed-up}} ↑ \\
        \midrule

        \multirow{2}{*}{\textbf{MathVista-mini}} &
        {\color{vanillacol}MMaDA} &
        32.90 & 1.00$\times$ \\
        & \cellcolor{oursbg1}{\textbf{Ours}}  &
        \cellcolor{oursbg1}{\textbf{36.00}} {\scriptsize\textcolor{inc}{(+3.10)}} &
        \cellcolor{oursbg1}{\textbf{2.37$\times$}} \\
        \midrule

        \multirow{2}{*}{\textbf{MATH-Vision}} &
        {\color{vanillacol}MMaDA} &
        9.44 & 1.00$\times$ \\
        & \cellcolor{oursbg1}{\textbf{Ours}} &
        \cellcolor{oursbg1}{\textbf{10.00}} {\scriptsize\textcolor{inc}{(+0.56)}} &
        \cellcolor{oursbg1}{\textbf{2.16$\times$}} \\
        \midrule

        \multirow{2}{*}{\textbf{ScienceQA}} &
        {\color{vanillacol}MMaDA} &
        48.88 & 1.00$\times$ \\
        & \cellcolor{oursbg1}{\textbf{Ours}} &
        \cellcolor{oursbg1}{\textbf{49.93}} {\scriptsize\textcolor{inc}{(+1.15)}} &
        \cellcolor{oursbg1}{\textbf{2.70$\times$}} \\
        \midrule

        \multirow{2}{*}{\textbf{GQA}} &
        {\color{vanillacol}MMaDA} &
        50.52 & 1.00$\times$ \\
        & \cellcolor{oursbg1}{\textbf{Ours}} &
        \cellcolor{oursbg1}{\textbf{50.58}} {\scriptsize\textcolor{inc}{(+0.06)}} &
        \cellcolor{oursbg1}{\textbf{16.40$\times$}} \\
        \midrule
        
        \multirow{2}{*}{\textbf{MME}} &
        {\color{vanillacol}MMaDA} &
        1074.97 & 1.00$\times$ \\
        & \cellcolor{oursbg1}{\textbf{Ours}} &
        \cellcolor{oursbg1}{\textbf{1075.72}} {\scriptsize\textcolor{inc}{(+0.75)}} &
        \cellcolor{oursbg1}{\textbf{15.98$\times$}} \\
        \bottomrule
    \end{tabular}
}
\vspace{-3mm}
\caption{Comparison of performance and inference speedup across different multi-modal benchmarks.}
\label{tab:vision_benchmarks}
\vspace{-6mm}
\end{table}

\begin{table}[ht]
    \centering

    \resizebox{\columnwidth}{!}{
    \begin{tabular}{l l c c}
        \toprule
        \textbf{Task} & \textbf{Method} &
        \emph{\textbf{Score}} ↑ &
        \emph{\textbf{Step}} ↓  \\
        \midrule

        \multirow{2}{*}{\textbf{AlpacaEval}} &
        {\color{vanillacol}DIFFA} &
        3.75 & 128 \\
        & \cellcolor{oursbg1}{\textbf{Ours}}  &
        \cellcolor{oursbg1}{\textbf{3.78}} {\scriptsize\textcolor{inc}{(+0.03)}} &
        \cellcolor{oursbg1}{\textbf{80.19}} {\scriptsize\textcolor{inc}{($-37\%$)}}  \\
        \midrule

        \multirow{2}{*}{\textbf{OpenBookQA}} &
        {\color{vanillacol}DIFFA} &
        36.50 & 128 \\
        & \cellcolor{oursbg1}{\textbf{Ours}} &
        \cellcolor{oursbg1}{\textbf{38.50}} {\scriptsize\textcolor{inc}{(+2.00)}} &
        \cellcolor{oursbg1}{\textbf{28.67}} {\scriptsize\textcolor{inc}{($-78\%$)}}  \\
        \midrule

        \multirow{2}{*}{\textbf{CommonEval}} &
        {\color{vanillacol}DIFFA} &
        2.90 & 128 \\
        & \cellcolor{oursbg1}{\textbf{Ours}} &
        \cellcolor{oursbg1}{\textbf{2.94}} {\scriptsize\textcolor{inc}{(+0.04)}} &
        \cellcolor{oursbg1}{\textbf{72.45}} {\scriptsize\textcolor{inc}{($-44\%$)}}  \\
        \midrule

        \multirow{2}{*}{\textbf{BBH}} &
        {\color{vanillacol}DIFFA} &
        53.00 & 128 \\
        & \cellcolor{oursbg1}{\textbf{Ours}} &
        \cellcolor{oursbg1}{\textbf{54.30}} {\scriptsize\textcolor{inc}{(+1.30)}} &
        \cellcolor{oursbg1}{\textbf{56.49}} {\scriptsize\textcolor{inc}{($-56\%$)}}  \\
        \midrule
        
        \multirow{2}{*}{\textbf{Wildvoice}} &
        {\color{vanillacol}DIFFA} &
        \textbf{2.80} & 128 \\
        & \cellcolor{oursbg1}{\textbf{Ours}} &
        \cellcolor{oursbg1}{2.76} {\scriptsize\textcolor{dec}{(-0.04)}} &
        \cellcolor{oursbg1}{\textbf{80.52}} {\scriptsize\textcolor{inc}{($-37\%$)}} \\
        \bottomrule
    \end{tabular}
}
\vspace{-3mm}
\caption{Evaluation results of DIFFA on VoiceBench.}
\label{tab:audio}
\vspace{-5mm}
\end{table}

\noindent\textbf{Evaluation Metrics.}
For each benchmark, we employ the standard performance evaluation metrics specified for that dataset. Additionally, to evaluate inference efficiency, we report both the decoding steps and latency (speed-up) as measures of computational speed. A reduction in decoding steps directly indicates the mitigation of redundant computations inherent in traditional dLLMs, thereby corresponding to decreased latency~\cite{wang2025creditdecoding,chen2025beyond,lu2025adablock}.



\subsection{Main Results}
\paragraph{Language Domain.}
We present the evaluation results for LLaDA-8B-Instruct and LLaDA-1.5 across seven benchmarks in Tables~\ref{tab:results-instruct} and \ref{tab:results-llada1.5}, comparing baseline approaches,  recent advanced decoding strategies, and our proposed AHD. Overall, our method consistently surpasses both naive decoding and existing advanced strategies in terms of both performance and inference efficiency. Notably, most advanced decoding methods achieve inference acceleration at the expense of model performance. Specifically, in rare cases where performance does improve, the gains are marginal (typically within 1\%), falling within the expected variation and not reflecting a significant enhancement. In contrast, our method achieves consistent performance improvements across all seven benchmarks while simultaneously accelerating inference. For instance, on the BBH benchmark, LLaDA achieves a performance gain of +3.67 points while reducing decoding steps by 80\%.



\noindent\textbf{Multimodal Domain.}
In addition to evaluating text-only models, we further assess multimodal understanding and reasoning. As shown in Table~\ref{tab:vision_benchmarks}, we present both performance metrics and inference efficiency (latency/speed-up) for MMaDA’s naive decoding and AHD across five benchmarks, where the latency (speed-up) of naive decoding is normalized to 1. Overall, AHD demonstrates consistent and stable improvements in both performance and inference efficiency across all five benchmarks. Notably, in multimodal mathematical reasoning, AHD achieves a +3.1 performance gain and 2.37× speedup on MathVista, underscoring its applicability to a wide range of multimodal tasks.



\noindent\textbf{Audio Domain.}
We utilize DIFFA to evaluate semantic audio understanding. As shown in Table~\ref{tab:audio}, our method outperforms DIFFA’s naive decoding on all five tasks in VoiceBench. For example, on OpenBookQA, AHD achieves a +2.0 performance gain while reducing decoding steps by 78\%. Although there is a slight performance decrease on Wildvoice, it is negligible compared with the improvement in inference efficiency.


\noindent\textbf{Scalability.}
Previous studies have primarily focused on short-sequence generation, typically restricting output length to 256 or 512 tokens and largely overlooking long-sequence scenarios. To evaluate the robustness of our method under extended lengths, we conduct experiments with outputs up to 1024 tokens. As presented in Table~\ref{tab:genlen_long_multi}, our approach consistently demonstrates superior performance even with increased generation lengths.

Intuitively, reducing decoding steps is typically perceived as inevitably leading to performance degradation. However, AHD primarily eliminates redundant iterations that have shown minimal variation or even introduced noise, while maintaining sufficient iterative capacity for tokens still undergoing meaningful evolutionary updates. Therefore, our method significantly reduces decoding steps while further improving the inference coherence and generation quality.




\noindent\textbf{Computational Complexity.}
When computing the anchored consistency score, there is some additional GPU memory usage and inference time. However, as shown in Figure~\ref{fig:ablation_hyperparameters} (a), this overhead is negligible compared with the total inference cost of dLLM, since the logits for all tokens have already been computed (see Appendix~\ref{related work}).


\subsection{Ablation Study}
In this section, we conduct comprehensive ablation experiments using LLaDA‑8B‑Instruct. More detailed experimental settings and results are provided in Appendix~\ref{sec:appendix-exper-detils} and~\ref{appendix_res}.



\begin{figure*}[t]
  \centering
  \includegraphics[width=\textwidth]{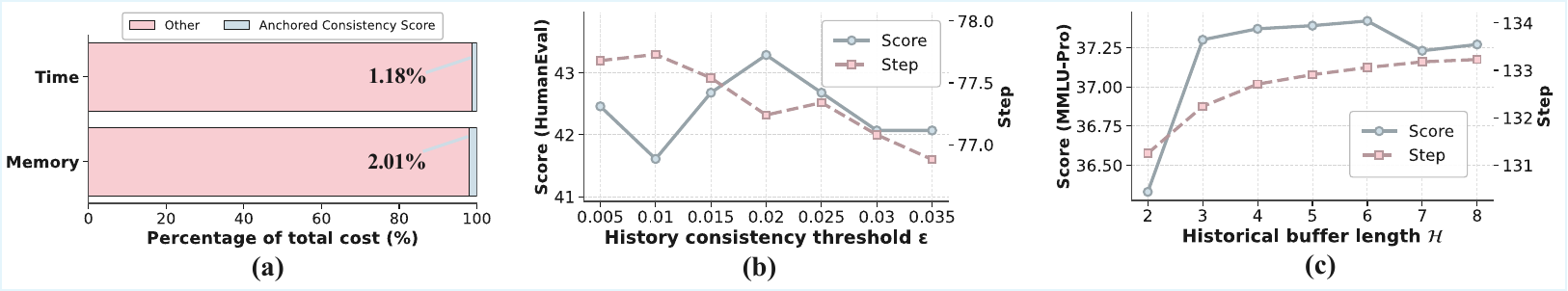} 
  \captionsetup{skip=1pt}
  \caption {(a) Computational complexity of related operations; (b) Ablation study on \(\varepsilon\); (c) Ablation study on \(\mathcal{H}\).}
  \vspace{-5mm}
  \label{fig:ablation_hyperparameters}
\end{figure*}

\noindent\textbf{Effect of Different Generation Length.}
To investigate the effect of generation length, we fix the block length at 32 and evaluate model performance with generation lengths of 128, 256, and 512. As summarized in Table~\ref{tab:genlen_mbpp_mmlu_bbh}, AHD consistently outperforms vanilla decoding in terms of both performance and inference efficiency across all generation length configurations. 



\begin{table}[t]
    \centering

    \resizebox{1\columnwidth}{!}{%
    \begin{tabular}{l
                    c c  
                    c c} 
        \toprule
        \multirow{2}{*}{\textbf{Method}} &
        \multicolumn{2}{c}{\textbf{BBH}} &
        \multicolumn{2}{c}{\textbf{HumanEval}} \\
        \cmidrule(lr){2-3}\cmidrule(lr){4-5}
        &
        \emph{Score}$\uparrow$ & \emph{Step}$\downarrow$ &
        \emph{Score}$\uparrow$ & \emph{Step}$\downarrow$ \\
        \midrule

        {\color{vanillacol}LLaDA} &
        53.28 & 1024 &
        43.90 & 1024 \\

        PC-sampler &
        54.03 & 1024 &
        42.07 & 1024 \\

        Fast-dLLM &
        53.68 & 65.54 &
        42.68 & 184.79 \\

        KLASS &
        53.42 & 122.66 &
        43.90 & 309.97 \\

        Saber &
        52.80 & 213.02 &
        43.90 & 210.43 \\

        \rowcolor{oursbg}\textbf{Ours} &
        \textbf{59.82} {\scriptsize\textcolor{inc}{(+6.54)}} & \textbf{58.25} {\scriptsize\textcolor{inc}{($-94\%$)}}  &
        \textbf{48.78} {\scriptsize\textcolor{inc}{(+4.88)}} & \textbf{146.48} {\scriptsize\textcolor{inc}{($-86\%$)}} \\
        \bottomrule
    \end{tabular}
    }
\vspace{-3mm}
\caption{Long-sequence results of LLaDA-8B-Instruct on two benchmarks (generation length 1024).}
\vspace{-2mm}
\label{tab:genlen_long_multi}
\end{table}

\begin{table}[t]
    \centering
    \footnotesize
    \setlength{\tabcolsep}{3pt}
    \renewcommand{\arraystretch}{1.05}

    \resizebox{\columnwidth}{!}{
    \begin{tabular}{l l c c c c c c}
        \toprule
        \multirow{2}{*}{\textbf{Task}} & \multirow{2}{*}{\textbf{Method}} &
        \multicolumn{2}{c}{\textbf{128}} &
        \multicolumn{2}{c}{\textbf{256}} &
        \multicolumn{2}{c}{\textbf{512}} \\
        \cmidrule(lr){3-4}\cmidrule(lr){5-6}\cmidrule(lr){7-8}
        & &
        \emph{Score}$\uparrow$ & \emph{Step}$\downarrow$ &
        \emph{Score}$\uparrow$ & \emph{Step}$\downarrow$ &
        \emph{Score}$\uparrow$ & \emph{Step}$\downarrow$ \\
        \midrule

        \multirow{2}{*}{\textbf{MBPP}} &
        {\color{vanillacol}LLaDA} &
        39.00 & 128 &
        29.20 & 256 &
        16.20 & 512 \\
        & \cellcolor{oursbg2}\textbf{Ours} &
        \cellcolor{oursbg2}\textbf{40.40} & \cellcolor{oursbg2}\textbf{47.65} &
        \cellcolor{oursbg2}\textbf{31.20} & \cellcolor{oursbg2}\textbf{66.03} &
        \cellcolor{oursbg2}\textbf{16.40} & \cellcolor{oursbg2}\textbf{116.75} \\
        \midrule

        \multirow{2}{*}{\textbf{MMLU-Pro}} &
        {\color{vanillacol}LLaDA} &
        37.42 & 128 &
        35.57 & 256 &
        37.46 & 512 \\
        & \cellcolor{oursbg2}\textbf{Ours} &
        \cellcolor{oursbg2}\textbf{38.12} & \cellcolor{oursbg2}\textbf{65.51} &
        \cellcolor{oursbg2}\textbf{37.42} & \cellcolor{oursbg2}\textbf{133.06} &
        \cellcolor{oursbg2}\textbf{39.00} & \cellcolor{oursbg2}\textbf{197.20} \\
        \midrule

        \multirow{2}{*}{\textbf{BBH}} &
        {\color{vanillacol}LLaDA} &
        45.71 & 128 &
        53.11 & 256 &
        53.14 & 512 \\
        & \cellcolor{oursbg2}\textbf{Ours} &
        \cellcolor{oursbg2}\textbf{50.07} & \cellcolor{oursbg2}\textbf{33.11} &
        \cellcolor{oursbg2}\textbf{55.69} & \cellcolor{oursbg2}\textbf{52.62} &
        \cellcolor{oursbg2}\textbf{59.05} & \cellcolor{oursbg2}\textbf{60.68} \\
        \bottomrule
    \end{tabular}
}
\vspace{-3mm}
\caption{Performance and inference speed on three benchmarks at different generation lengths.}
\label{tab:genlen_mbpp_mmlu_bbh}
\vspace{-4mm}
\end{table}

\noindent\textbf{Effect of Different Block Length.}
To assess the influence of block length, the generation length is fixed at 256, and AHD is applied with four block configurations: 16, 32, 64, and 128. As reported in Table~\ref{tab:blocklen_by_task}, AHD maintains superior task performance and inference efficiency over vanilla decoding for all block lengths, indicating strong robustness.
Notably, this robustness  extends to multimodal configurations, as evidenced in Table~\ref{tab:mutilmodel_ablation}.



\begin{table}[t]
    \centering
    \footnotesize
    \setlength{\tabcolsep}{3pt}
    \renewcommand{\arraystretch}{1.05}

    \resizebox{\columnwidth}{!}{
    \begin{tabular}{l l c c c c c c}
        \toprule
        \textbf{Block} & \multirow{2}{*}{\textbf{Method}} &
        \multicolumn{2}{c}{\textbf{MBPP}} &
        \multicolumn{2}{c}{\textbf{MMLU-Pro}} &
        \multicolumn{2}{c}{\textbf{BBH}} \\
        \cmidrule(lr){3-4}\cmidrule(lr){5-6}\cmidrule(lr){7-8}
        \textbf{length} & &
        \emph{Score}$\uparrow$ & \emph{Step}$\downarrow$ &
        \emph{Score}$\uparrow$ & \emph{Step}$\downarrow$ &
        \emph{Score}$\uparrow$ & \emph{Step}$\downarrow$ \\
        \midrule

        \multirow{2}{*}{\textbf{16}} &
        {\color{vanillacol}LLaDA} &
        28.60 & 256 &
        33.74 & 256 &
        52.51 & 256 \\
        & \cellcolor{oursbg2}\textbf{Ours} &
        \cellcolor{oursbg2}\textbf{30.80} & \cellcolor{oursbg2}\textbf{72.73} &
        \cellcolor{oursbg2}\textbf{37.36} & \cellcolor{oursbg2}\textbf{137.02} &
        \cellcolor{oursbg2}\textbf{56.12} & \cellcolor{oursbg2}\textbf{59.88} \\
        \midrule

        \multirow{2}{*}{\textbf{32}} &
        {\color{vanillacol}LLaDA} &
        29.20 & 256 &
        35.57 & 256 &
        53.11 & 256 \\
        & \cellcolor{oursbg2}\textbf{Ours} &
        \cellcolor{oursbg2}\textbf{31.20} & \cellcolor{oursbg2}\textbf{66.03} &
        \cellcolor{oursbg2}\textbf{37.42} & \cellcolor{oursbg2}\textbf{133.06} &
        \cellcolor{oursbg2}\textbf{55.69} & \cellcolor{oursbg2}\textbf{52.62} \\
        \midrule

        \multirow{2}{*}{\textbf{64}} &
        {\color{vanillacol}LLaDA} &
        27.40 & 256 &
        36.67 & 256 &
        54.32 & 256 \\
        & \cellcolor{oursbg2}\textbf{Ours} &
        \cellcolor{oursbg2}\textbf{28.80} & \cellcolor{oursbg2}\textbf{63.50} &
        \cellcolor{oursbg2}\textbf{38.02} & \cellcolor{oursbg2}\textbf{131.54} &
        \cellcolor{oursbg2}\textbf{55.57} & \cellcolor{oursbg2}\textbf{48.93} \\
        \midrule

        \multirow{2}{*}{\textbf{128}} &
        {\color{vanillacol}LLaDA} &
        \textbf{26.40} & 256 &
        36.67 & 256 &
        53.46 & 256 \\
        & \cellcolor{oursbg2}\textbf{Ours} &
        \cellcolor{oursbg2}\textbf{26.40} & \cellcolor{oursbg2}\textbf{59.47} &
        \cellcolor{oursbg2}\textbf{37.84} & \cellcolor{oursbg2}\textbf{132.14} &
        \cellcolor{oursbg2}\textbf{53.49} & \cellcolor{oursbg2}\textbf{48.08} \\
        \bottomrule
    \end{tabular}
    }
\vspace{-3mm}
\caption{Performance and inference speed on three benchmarks at different block lengths.}
\vspace{-2mm}
\label{tab:blocklen_by_task}
\end{table}

\begin{table}[ht]
    \centering

    \resizebox{\columnwidth}{!}{
    \begin{tabular}{l c c c c c c c c}
        \toprule
        \multirow{2}{*}{\textbf{Method}} &
        \multicolumn{2}{c}{\textbf{16}} &
        \multicolumn{2}{c}{\textbf{32}} &
        \multicolumn{2}{c}{\textbf{64}} &
        \multicolumn{2}{c}{\textbf{128}} \\
        \cmidrule(lr){2-3}\cmidrule(lr){4-5}\cmidrule(lr){6-7}\cmidrule(lr){8-9}
        &
        \emph{Score}$\uparrow$ & \emph{Speed Up}$\uparrow$ &
        \emph{Score}$\uparrow$ & \emph{Speed Up}$\uparrow$ &
        \emph{Score}$\uparrow$ & \emph{Speed Up}$\uparrow$ &
        \emph{Score}$\uparrow$ & \emph{Speed Up}$\uparrow$ \\
        \midrule

        {\color{vanillacol}MMaDA} &
        31.3 & 1.00$\times$ &
        32.1 & 1.00$\times$ &
        32.9 & 1.00$\times$ &
        32.5 & 1.00$\times$ \\
        \cellcolor{oursbg2}\textbf{Ours} &
        \cellcolor{oursbg2}\textbf{33.9} & \cellcolor{oursbg2}\textbf{2.01$\times$} &
        \cellcolor{oursbg2}\textbf{34.0} & \cellcolor{oursbg2}\textbf{2.08$\times$} &
        \cellcolor{oursbg2}\textbf{36.0} & \cellcolor{oursbg2}\textbf{2.37$\times$} &
        \cellcolor{oursbg2}\textbf{32.9} & \cellcolor{oursbg2}\textbf{2.04$\times$} \\
        \bottomrule
    \end{tabular}
}
\vspace{-3mm}
\caption{Performance and inference speed on Mathvista-mini benchmark at different block lengths.}
\label{tab:mutilmodel_ablation}
\vspace{-2mm}
\end{table}

\begin{table}[!ht]
    \centering

    \resizebox{\columnwidth}{!}{
    \begin{tabular}{l c c c c c c c c c c}
        \toprule
        \multirow{2}{*}{\textbf{Method}} &
        \multicolumn{2}{c}{\textbf{temperature=0}} &
        \multicolumn{2}{c}{\textbf{temperature=0.1}} &
        \multicolumn{2}{c}{\textbf{temperature=0.4}} &
        \multicolumn{2}{c}{\textbf{temperature=0.7}} &
        \multicolumn{2}{c}{\textbf{temperature=1}} \\
        \cmidrule(lr){2-3}\cmidrule(lr){4-5}\cmidrule(lr){6-7}\cmidrule(lr){8-9}\cmidrule(lr){10-11}
        &
        \emph{Score}$\uparrow$ & \emph{Step}$\downarrow$ &
        \emph{Score}$\uparrow$ & \emph{Step}$\downarrow$ &
        \emph{Score}$\uparrow$ & \emph{Step}$\downarrow$ &
        \emph{Score}$\uparrow$ & \emph{Step}$\downarrow$ &
        \emph{Score}$\uparrow$ & \emph{Step}$\downarrow$ \\
        \midrule

        {\color{vanillacol}LLaDA} &
        35.57 & 256 &
        35.86 & 256 &
        35.80 & 256 &
        35.73 & 256 &
        35.20 & 256 \\

        \cellcolor{oursbg2}\textbf{Ours} &
        \cellcolor{oursbg2}\textbf{37.42} & \cellcolor{oursbg2}\textbf{133.06} &
        \cellcolor{oursbg2}\textbf{37.99} & \cellcolor{oursbg2}\textbf{132.77} &
        \cellcolor{oursbg2}\textbf{37.46} & \cellcolor{oursbg2}\textbf{132.08} &
        \cellcolor{oursbg2}\textbf{37.22} & \cellcolor{oursbg2}\textbf{131.06} &
        \cellcolor{oursbg2}\textbf{36.93} & \cellcolor{oursbg2}\textbf{131.62} \\
        \bottomrule
    \end{tabular}
}
\vspace{-3mm}
\caption{Performance and decoding steps on MMLU-Pro under different temperature coefficients.}
\label{tab:temp_mmlu_pro}
\vspace{-3mm}
\end{table}

\noindent\textbf{Effect of Historical Consistency Threshold \(\varepsilon\).}
We conducted ablation experiments to analyze the impact of the \(\varepsilon\) on performance and decoding steps. As illustrated in Figure~\ref{fig:ablation_hyperparameters} (b), small values of \(\varepsilon\) prevent the early release of cross-block stable tokens, resulting in increased fluctuations and degraded performance~\cite{wang2025time}.
Conversely, excessively large values of \(\varepsilon\) result in the premature release of numerous unstable tokens across blocks, similarly impairing performance. In addition, the decoding steps decrease monotonically with increasing \(\varepsilon\). Notably, when \(\varepsilon = 0.02\), we achieve a balance between inference efficiency and performance,  representing an optimal trade-off.



\noindent\textbf{Effect of Historical Buffer Length \( \mathcal{H} \).}
Our proposed method can achieve a trade‑off between performance and inference efficiency by adjusting the historical buffer length. As shown in Figure~\ref{fig:ablation_hyperparameters} (c), a significantly small  \(\mathcal{H}\), such as 2, leads to a substantial performance degradation. This decline occurs because a short monitoring history fails to capture the true \emph{\textbf{absolute stability trend}}, as noted in \textbf{core insights 1 and 2}. Conversely, we observe optimal performance when \(\mathcal{H} = 6\), indicating an effective compromise between capturing historical trends and maintaining computational efficiency.



\noindent\textbf{Robustness to Temperature Coefficients.}
In the main experiments, we use greedy decoding to ensure reproducibility. Additionally, we investigate the impact of temperature on AHD. Table~\ref{tab:temp_mmlu_pro} shows AHD exhibits strong robustness across varying temperature settings, while consistently delivering notable gains in accuracy and decoding efficiency.


\section{Conclusion}
In this work, we propose Anchor-based History-stable Decoding (AHD). Its core idea is to use dynamic anchors to monitor the historical trajectory in real time, capture the absolute stability trend at an early stage, and thereby unlock stable tokens across blocks in advance. AHD substantially reduces the decoding steps while consistently improving performance across multiple benchmarks in the language, multimodal, and audio domains. This offers a novel perspective for the development of future advanced decoding strategies.


\section*{Limitations}
In this section, we discuss several limitations of our work.
(1) Although AHD is a training‑free, plug‑and‑play method, its parameters vary slightly across tasks, which increases the adaptation cost for new scenarios.
(2) Our experiments mainly focus on models of around 8B parameters, and have not yet validated effectiveness on larger models such as 72B or 256B. With more computational resources, we plan to explore longer sequences and larger‑scale models.
(3) Although the additional computation introduced by AHD is almost negligible, we further optimize the related operations, such as vectorization, to further improve inference efficiency.



\bibliography{custom}

\begin{thebibliography}{58}
\providecommand{\natexlab}[1]{#1}

\bibitem[{Austin et~al.(2021{\natexlab{a}})Austin, Johnson, Ho, Tarlow, and Van Den~Berg}]{austin2021structured}
Jacob Austin, Daniel~D Johnson, Jonathan Ho, Daniel Tarlow, and Rianne Van Den~Berg. 2021{\natexlab{a}}.
\newblock \href {https://dl.acm.org/doi/10.5555/3540261.3541637} {Structured denoising diffusion models in discrete state-spaces}.
\newblock \emph{Advances in neural information processing systems}.

\bibitem[{Austin et~al.(2021{\natexlab{b}})Austin, Odena, Nye, Bosma, Michalewski, Dohan, Jiang, Cai, Terry, Le, and Sutton}]{austin2021program}
Jacob Austin, Augustus Odena, Maxwell Nye, Maarten Bosma, Henryk Michalewski, David Dohan, Ellen Jiang, Carrie Cai, Michael Terry, Quoc Le, and Charles Sutton. 2021{\natexlab{b}}.
\newblock \href {https://arxiv.org/abs/2108.07732} {Program synthesis with large language models}.
\newblock \emph{arXiv preprint arXiv:2108.07732}.

\bibitem[{Ben-Hamu et~al.(2025)Ben-Hamu, Gat, Severo, Nolte, and Karrer}]{ben2025accelerated}
Heli Ben-Hamu, Itai Gat, Daniel Severo, Niklas Nolte, and Brian Karrer. 2025.
\newblock \href {https://arxiv.org/abs/2505.24857} {Accelerated sampling from masked diffusion models via entropy bounded unmasking}.
\newblock \emph{arXiv preprint arXiv:2505.24857}.

\bibitem[{Berglund et~al.(2024)Berglund, Tong, Kaufmann, Balesni, Stickland, Korbak, and Evans}]{berglund2023reversal}
Lukas Berglund, Meg Tong, Max Kaufmann, Mikita Balesni, Asa~Cooper Stickland, Tomasz Korbak, and Owain Evans. 2024.
\newblock \href {https://arxiv.org/abs/2309.12288} {The reversal curse: Llms trained on "a is b" fail to learn "b is a"}.
\newblock \emph{arXiv preprint arXiv:2309.12288}.

\bibitem[{Bie et~al.(2025)Bie, Cao, Chen, Du, Gong, Gong, Gu, Hu, Huang, Lan, Li, Li, Li, Li, Liu, Liu, Lu, Lu, Ma, Tan, Wei, Wen, Xing, Zhang, Zhao, Zheng, Zhou, Zhou, Zhou, Zhu, and Zhuang}]{bie2025llada2}
Tiwei Bie, Maosong Cao, Kun Chen, Lun Du, Mingliang Gong, Zhuochen Gong, Yanmei Gu, Jiaqi Hu, Zenan Huang, Zhenzhong Lan, Chengxi Li, Chongxuan Li, Jianguo Li, Zehuan Li, Huabin Liu, Lin Liu, Guoshan Lu, Xiaocheng Lu, Yuxin Ma, and 12 others. 2025.
\newblock \href {https://arxiv.org/abs/2512.15745} {Llada2.0: Scaling up diffusion language models to 100b}.
\newblock \emph{arXiv preprint arXiv:2512.15745}.

\bibitem[{Campbell et~al.(2022)Campbell, Benton, De~Bortoli, Rainforth, Deligiannidis, and Doucet}]{campbell2022continuous}
Andrew Campbell, Joe Benton, Valentin De~Bortoli, Thomas Rainforth, George Deligiannidis, and Arnaud Doucet. 2022.
\newblock \href {https://dl.acm.org/doi/10.5555/3600270.3602319} {A continuous time framework for discrete denoising models}.
\newblock \emph{Advances in Neural Information Processing Systems}.

\bibitem[{Chen et~al.(2023)Chen, Zhang, Li, Smola, and Yang}]{chen2023cheaper}
Jiaao Chen, Aston Zhang, Mu~Li, Alex Smola, and Diyi Yang. 2023.
\newblock \href {https://aclanthology.org/2023.emnlp-main.289/} {A cheaper and better diffusion language model with soft-masked noise}.
\newblock In \emph{Proceedings of the 2023 Conference on Empirical Methods in Natural Language Processing}.

\bibitem[{Chen et~al.(2025{\natexlab{a}})Chen, Liu, Tao, Liu, Fu, Zhang, Tu, Kong, Liu, and Li}]{chen2025beyond}
Kecheng Chen, Ziru Liu, Xijia Tao, Hui Liu, Xinyu Fu, Suiyun Zhang, Dandan Tu, Lingpeng Kong, Rui Liu, and Haoliang Li. 2025{\natexlab{a}}.
\newblock \href {https://arxiv.org/abs/2512.02044} {Beyond confidence: Adaptive and coherent decoding for diffusion language models}.
\newblock \emph{arXiv preprint arXiv:2512.02044}.

\bibitem[{Chen et~al.(2021)Chen, Tworek, Jun, Yuan, de~Oliveira~Pinto, Kaplan, Edwards, Burda, Joseph, Brockman, Ray, Puri, Krueger, Petrov, Khlaaf, Sastry, Mishkin, Chan, Gray, Ryder, Pavlov, Power, Kaiser, Bavarian, Winter, Tillet, Such, Cummings, Plappert, Chantzis, Barnes, Herbert-Voss, Guss, Nichol, Paino, Tezak, Tang, Babuschkin, Balaji, Jain, Saunders, Hesse, Carr, Leike, Achiam, Misra, Morikawa, Radford, Knight, Brundage, Murati, Mayer, Welinder, McGrew, Amodei, McCandlish, Sutskever, and Zaremba}]{chen2021codex}
Mark Chen, Jerry Tworek, Heewoo Jun, Qiming Yuan, Henrique~Ponde de~Oliveira~Pinto, Jared Kaplan, Harri Edwards, Yuri Burda, Nicholas Joseph, Greg Brockman, Alex Ray, Raul Puri, Gretchen Krueger, Michael Petrov, Heidy Khlaaf, Girish Sastry, Pamela Mishkin, Brooke Chan, Scott Gray, and 39 others. 2021.
\newblock \href {https://arxiv.org/abs/2107.03374} {Evaluating large language models trained on code}.
\newblock \emph{arXiv preprint arXiv:2107.03374}.

\bibitem[{Chen et~al.(2025{\natexlab{b}})Chen, Huang, Guo, Wei, He, Zhang, Li, and Chen}]{chen2025dpad}
Xinhua Chen, Sitao Huang, Cong Guo, Chiyue Wei, Yintao He, Jianyi Zhang, Hai~"Helen" Li, and Yiran Chen. 2025{\natexlab{b}}.
\newblock \href {https://arxiv.org/abs/2508.14148} {Dpad: Efficient diffusion language models with suffix dropout}.
\newblock \emph{arXiv preprint arXiv:2508.14148}.

\bibitem[{Chen et~al.(2024)Chen, Yue, Zhang, Gao, Tan, and Li}]{chen2024voicebench}
Yiming Chen, Xianghu Yue, Chen Zhang, Xiaoxue Gao, Robby~T. Tan, and Haizhou Li. 2024.
\newblock \href {https://arxiv.org/abs/2410.17196} {Voicebench: Benchmarking llm-based voice assistants}.
\newblock \emph{arXiv preprint arXiv:2410.17196}.

\bibitem[{Cobbe et~al.(2021)Cobbe, Kosaraju, Bavarian, Chen, Jun, Kaiser, Plappert, Tworek, Hilton, Nakano, Hesse, and Schulman}]{cobbe2021training}
Karl Cobbe, Vineet Kosaraju, Mohammad Bavarian, Mark Chen, Heewoo Jun, Lukasz Kaiser, Matthias Plappert, Jerry Tworek, Jacob Hilton, Reiichiro Nakano, Christopher Hesse, and John Schulman. 2021.
\newblock \href {https://arxiv.org/abs/2110.14168} {Training verifiers to solve math word problems}.
\newblock \emph{arXiv preprint arXiv:2110.14168}.

\bibitem[{Dong et~al.(2025)Dong, Ma, Jiang, Fan, Qian, Li, Xiao, Jin, Cao, Li, Huang, Li, and Li}]{dong2025saber}
Yihong Dong, Zhaoyu Ma, Xue Jiang, Zhiyuan Fan, Jiaru Qian, Yongmin Li, Jianha Xiao, Zhi Jin, Rongyu Cao, Binhua Li, Fei Huang, Yongbin Li, and Ge~Li. 2025.
\newblock \href {https://arxiv.org/abs/2510.18165} {Saber: An efficient sampling with adaptive acceleration and backtracking enhanced remasking for diffusion language model}.
\newblock \emph{arXiv preprint arXiv:2510.18165}.

\bibitem[{Fu et~al.(2025)Fu, Chen, Shen, Qin, Zhang, Lin, Yang, Zheng, Li, Sun, Wu, Ji, Shan, and He}]{fu2023mme}
Chaoyou Fu, Peixian Chen, Yunhang Shen, Yulei Qin, Mengdan Zhang, Xu~Lin, Jinrui Yang, Xiawu Zheng, Ke~Li, Xing Sun, Yunsheng Wu, Rongrong Ji, Caifeng Shan, and Ran He. 2025.
\newblock \href {https://arxiv.org/abs/2306.13394} {Mme: A comprehensive evaluation benchmark for multimodal large language models}.
\newblock \emph{arXiv preprint arXiv:2306.13394}.

\bibitem[{Gong et~al.(2025)Gong, Agarwal, Zhang, Ye, Zheng, Li, An, Zhao, Bi, Han, Peng, and Kong}]{gong2024scaling}
Shansan Gong, Shivam Agarwal, Yizhe Zhang, Jiacheng Ye, Lin Zheng, Mukai Li, Chenxin An, Peilin Zhao, Wei Bi, Jiawei Han, Hao Peng, and Lingpeng Kong. 2025.
\newblock \href {https://arxiv.org/abs/2410.17891} {Scaling diffusion language models via adaptation from autoregressive models}.
\newblock \emph{arXiv preprint arXiv:2410.17891}.

\bibitem[{Hendrycks et~al.(2021)Hendrycks, Burns, Kadavath, Arora, Basart, Tang, Song, and Steinhardt}]{hendrycks2021measuring}
Dan Hendrycks, Collin Burns, Saurav Kadavath, Akul Arora, Steven Basart, Eric Tang, Dawn Song, and Jacob Steinhardt. 2021.
\newblock \href {https://arxiv.org/abs/2103.03874} {Measuring mathematical problem solving with the math dataset}.
\newblock \emph{arXiv preprint arXiv:2103.03874}.

\bibitem[{Ho et~al.(2020)Ho, Jain, and Abbeel}]{ho2020denoising}
Jonathan Ho, Ajay Jain, and Pieter Abbeel. 2020.
\newblock \href {https://dl.acm.org/doi/abs/10.5555/3495724.3496298} {Denoising diffusion probabilistic models}.
\newblock \emph{Advances in Neural Information Processing Systems}.

\bibitem[{Hong et~al.(2025)Hong, Yu, Ye, Huang, Zheng, Zhang, Wang, and Yao}]{hong2025wide}
Feng Hong, Geng Yu, Yushi Ye, Haicheng Huang, Huangjie Zheng, Ya~Zhang, Yanfeng Wang, and Jiangchao Yao. 2025.
\newblock \href {https://arxiv.org/abs/2507.18578} {Wide-in, narrow-out: Revokable decoding for efficient and effective dllms}.
\newblock \emph{arXiv preprint arXiv:2507.18578}.

\bibitem[{Huang et~al.(2025{\natexlab{a}})Huang, Zhang, Yang, Huang, Qi, Liu, and Zhang}]{huang2025mask}
Jianuo Huang, Yaojie Zhang, Yicun Yang, Benhao Huang, Biqing Qi, Dongrui Liu, and Linfeng Zhang. 2025{\natexlab{a}}.
\newblock \href {https://arxiv.org/abs/2510.09309} {Mask tokens as prophet: Fine-grained cache eviction for efficient dllm inference}.
\newblock \emph{arXiv preprint arXiv:2510.09309}.

\bibitem[{Huang et~al.(2025{\natexlab{b}})Huang, Liu, Liu, Yan, Wang, Chen, and Xiao}]{huang2025pc}
Pengcheng Huang, Shuhao Liu, Zhenghao Liu, Yukun Yan, Shuo Wang, Zulong Chen, and Tong Xiao. 2025{\natexlab{b}}.
\newblock \href {https://arxiv.org/abs/2508.13021v1} {Pc-sampler: Position-aware calibration of decoding bias in masked diffusion models}.
\newblock \emph{arXiv preprint arXiv:2508.13021}.

\bibitem[{Hudson and Manning(2019)}]{hudson2019gqa}
Drew~A Hudson and Christopher~D Manning. 2019.
\newblock \href {https://openaccess.thecvf.com/content_CVPR_2019/html/Hudson_GQA_A_New_Dataset_for_Real-World_Visual_Reasoning_and_Compositional_CVPR_2019_paper.html} {Gqa: A new dataset for real-world visual reasoning and compositional question answering}.
\newblock In \emph{Proceedings of the IEEE/CVF conference on computer vision and pattern recognition}.

\bibitem[{Kim et~al.(2026)Kim, Hong, Jung, Park, and Yun}]{kim2025klass}
Seo~Hyun Kim, Sunwoo Hong, Hojung Jung, Youngrok Park, and Se-Young Yun. 2026.
\newblock \href {https://arxiv.org/abs/2511.05664} {Klass: Kl-guided fast inference in masked diffusion models}.
\newblock \emph{arXiv preprint arXiv:2511.05664}.

\bibitem[{Kong et~al.(2025)Kong, Zhang, Liu, Wu, Tian, W., and Zhou}]{kong2025accelerating}
Fanheng Kong, Jingyuan Zhang, Yahui Liu, Zirui Wu, Yu~Tian, Victoria W., and Guorui Zhou. 2025.
\newblock \href {https://arxiv.org/abs/2510.07081} {Accelerating diffusion llm inference via local determinism propagation}.
\newblock \emph{arXiv preprint arXiv:2510.07081}.

\bibitem[{Li et~al.(2026)Li, Zhou, Muhtar, Yin, Yan, Shen, Vosoughi, and Liu}]{li2025diffusion}
Pengxiang Li, Yefan Zhou, Dilxat Muhtar, Lu~Yin, Shilin Yan, Li~Shen, Soroush Vosoughi, and Shiwei Liu. 2026.
\newblock \href {https://arxiv.org/abs/2508.19982} {Diffusion language models know the answer before decoding}.
\newblock \emph{arXiv preprint arXiv:2508.19982}.

\bibitem[{Lin et~al.(2022)Lin, Hilton, and Evans}]{lin2022truthfulqa}
Stephanie Lin, Jacob Hilton, and Owain Evans. 2022.
\newblock \href {https://aclanthology.org/2022.acl-long.229/} {Truthfulqa: Measuring how models mimic human falsehoods}.
\newblock In \emph{Proceedings of the 60th annual meeting of the association for computational linguistics (volume 1: long papers)}.

\bibitem[{Liu et~al.(2025)Liu, Yang, Zhang, Chen, Zou, Wei, Wang, and Zhang}]{liu2025dllm}
Zhiyuan Liu, Yicun Yang, Yaojie Zhang, Junjie Chen, Chang Zou, Qingyuan Wei, Shaobo Wang, and Linfeng Zhang. 2025.
\newblock \href {https://arxiv.org/abs/2506.06295} {dllm-cache: Accelerating diffusion large language models with adaptive caching}.
\newblock \emph{arXiv preprint arXiv:2506.06295}.

\bibitem[{Lovelace et~al.(2023)Lovelace, Kishore, Wan, Shekhtman, and Weinberger}]{lovelace2023latent}
Justin Lovelace, Varsha Kishore, Chao Wan, Eliot Shekhtman, and Kilian~Q Weinberger. 2023.
\newblock \href {https://dl.acm.org/doi/10.5555/3666122.3668614} {Latent diffusion for language generation}.
\newblock \emph{Advances in Neural Information Processing Systems}.

\bibitem[{Lu et~al.(2026)Lu, Chen, Karashima, Wang, Fujiki, and Fan}]{lu2025adablock}
Guanxi Lu, Hao~Mark Chen, Yuto Karashima, Zhican Wang, Daichi Fujiki, and Hongxiang Fan. 2026.
\newblock \href {https://arxiv.org/abs/2509.26432} {Adablock-dllm: Semantic-aware diffusion llm inference via adaptive block size}.
\newblock \emph{arXiv preprint arXiv:2509.26432}.

\bibitem[{Lu et~al.(2024)Lu, Bansal, Xia, Liu, Li, Hajishirzi, Cheng, Chang, Galley, and Gao}]{lu2023mathvista}
Pan Lu, Hritik Bansal, Tony Xia, Jiacheng Liu, Chunyuan Li, Hannaneh Hajishirzi, Hao Cheng, Kai-Wei Chang, Michel Galley, and Jianfeng Gao. 2024.
\newblock \href {https://arxiv.org/abs/2310.02255} {Mathvista: Evaluating mathematical reasoning of foundation models in visual contexts}.
\newblock \emph{arXiv preprint arXiv:2310.02255}.

\bibitem[{Lu et~al.(2022)Lu, Mishra, Xia, Qiu, Chang, Zhu, Tafjord, Clark, and Kalyan}]{lu2022learn}
Pan Lu, Swaroop Mishra, Tony Xia, Liang Qiu, Kai-Wei Chang, Song-Chun Zhu, Oyvind Tafjord, Peter Clark, and Ashwin Kalyan. 2022.
\newblock \href {https://dl.acm.org/doi/abs/10.5555/3600270.3600452} {Learn to explain: Multimodal reasoning via thought chains for science question answering}.
\newblock \emph{Advances in Neural Information Processing Systems}.

\bibitem[{Ma et~al.(2025{\natexlab{a}})Ma, Yu, Fang, and Wang}]{ma2025dkv}
Xinyin Ma, Runpeng Yu, Gongfan Fang, and Xinchao Wang. 2025{\natexlab{a}}.
\newblock \href {https://arxiv.org/abs/2505.15781} {dkv-cache: The cache for diffusion language models}.
\newblock \emph{arXiv preprint arXiv:2505.15781}.

\bibitem[{Ma et~al.(2025{\natexlab{b}})Ma, Du, Wei, Chen, Xu, Wang, Feng, Lu, Liu, Qi, Zhang, Tao, Feng, Jiang, Xu, Huang, Zhuang, Xu, Hu, Lan, Zhao, Li, and Zheng}]{ma2025dinfer}
Yuxin Ma, Lun Du, Lanning Wei, Kun Chen, Qian Xu, Kangyu Wang, Guofeng Feng, Guoshan Lu, Lin Liu, Xiaojing Qi, Xinyuan Zhang, Zhen Tao, Haibo Feng, Ziyun Jiang, Ying Xu, Zenan Huang, Yihong Zhuang, Haokai Xu, Jiaqi Hu, and 4 others. 2025{\natexlab{b}}.
\newblock \href {https://arxiv.org/abs/2510.08666} {dinfer: An efficient inference framework for diffusion language models}.
\newblock \emph{arXiv preprint arXiv:2510.08666}.

\bibitem[{Melnik et~al.(2024)Melnik, Ljubljanac, Lu, Yan, Ren, and Ritter}]{melnik2024video}
Andrew Melnik, Michal Ljubljanac, Cong Lu, Qi~Yan, Weiming Ren, and Helge Ritter. 2024.
\newblock \href {https://arxiv.org/abs/2405.03150} {Video diffusion models: A survey}.
\newblock \emph{arXiv preprint arXiv:2405.03150}.

\bibitem[{Miao et~al.(2020)Miao, Liang, and Su}]{miao2020diverse}
Shen-Yun Miao, Chao-Chun Liang, and Keh-Yih Su. 2020.
\newblock \href {https://aclanthology.org/2020.acl-main.92/} {A diverse corpus for evaluating and developing english math word problem solvers}.
\newblock In \emph{Proceedings of the 58th annual meeting of the Association for Computational Linguistics}.

\bibitem[{Nie et~al.(2025)Nie, Zhu, You, Zhang, Ou, Hu, Zhou, Lin, Wen, and Li}]{nie2025large}
Shen Nie, Fengqi Zhu, Zebin You, Xiaolu Zhang, Jingyang Ou, Jun Hu, Jun Zhou, Yankai Lin, Ji-Rong Wen, and Chongxuan Li. 2025.
\newblock \href {https://arxiv.org/abs/2502.09992} {Large language diffusion models}.
\newblock \emph{arXiv preprint arXiv:2502.09992}.

\bibitem[{Ou et~al.(2026)Ou, Nie, Xue, Zhu, Sun, Li, and Li}]{ou2024your}
Jingyang Ou, Shen Nie, Kaiwen Xue, Fengqi Zhu, Jiacheng Sun, Zhenguo Li, and Chongxuan Li. 2026.
\newblock \href {https://arxiv.org/abs/2406.03736} {Your absorbing discrete diffusion secretly models the conditional distributions of clean data}.
\newblock \emph{arXiv preprint arXiv:2406.03736}.

\bibitem[{Rombach et~al.(2022)Rombach, Blattmann, Lorenz, Esser, and Ommer}]{rombach2022high}
Robin Rombach, Andreas Blattmann, Dominik Lorenz, Patrick Esser, and Bj{\"o}rn Ommer. 2022.
\newblock \href {https://openaccess.thecvf.com/content/CVPR2022/html/Rombach_High-Resolution_Image_Synthesis_With_Latent_Diffusion_Models_CVPR_2022_paper.html} {High-resolution image synthesis with latent diffusion models}.
\newblock In \emph{Proceedings of the IEEE/CVF conference on computer vision and pattern recognition}.

\bibitem[{Sahoo et~al.(2024)Sahoo, Arriola, Schiff, Gokaslan, Marroquin, Chiu, Rush, and Kuleshov}]{sahoo2024simple}
Subham~Sekhar Sahoo, Marianne Arriola, Yair Schiff, Aaron Gokaslan, Edgar Marroquin, Justin~T Chiu, Alexander Rush, and Volodymyr Kuleshov. 2024.
\newblock \href {https://dl.acm.org/doi/10.5555/3737916.3742051} {Simple and effective masked diffusion language models}.
\newblock \emph{Advances in Neural Information Processing Systems}.

\bibitem[{Shi et~al.(2024)Shi, Han, Wang, Doucet, and Titsias}]{shi2024simplified}
Jiaxin Shi, Kehang Han, Zhe Wang, Arnaud Doucet, and Michalis~K. Titsias. 2024.
\newblock \href {https://dl.acm.org/doi/10.5555/3737916.3741193} {Simplified and generalized masked diffusion for discrete data}.
\newblock \emph{Advances in Neural Information Processing Systems}.

\bibitem[{Song et~al.(2025)Song, Liu, Li, Liu, Huang, Guo, He, and Qiu}]{song2025sparse}
Yuerong Song, Xiaoran Liu, Ruixiao Li, Zhigeng Liu, Zengfeng Huang, Qipeng Guo, Ziwei He, and Xipeng Qiu. 2025.
\newblock \href {https://arxiv.org/abs/2508.02558} {Sparse-dllm: Accelerating diffusion llms with dynamic cache eviction}.
\newblock \emph{arXiv preprint arXiv:2508.02558}.

\bibitem[{Suzgun et~al.(2023)Suzgun, Scales, Schärli, Gehrmann, Tay, Chung, Chowdhery, Le, Chi, Zhou, and Wei}]{suzgun2023challenging}
Mirac Suzgun, Nathan Scales, Nathanael Schärli, Sebastian Gehrmann, Yi~Tay, Hyung~Won Chung, Aakanksha Chowdhery, Quoc~V. Le, Ed~H. Chi, Denny Zhou, and Jason Wei. 2023.
\newblock \href {https://aclanthology.org/2023.findings-acl.824/} {Challenging {BIG}-bench tasks and whether chain-of-thought can solve them}.
\newblock In \emph{Findings of the Association for Computational Linguistics: ACL 2023}.

\bibitem[{Wang et~al.(2025{\natexlab{a}})Wang, Jiang, Feng, Zhao, Liu, Li, Lan, and Lin}]{wang2025creditdecoding}
Kangyu Wang, Zhiyun Jiang, Haibo Feng, Weijia Zhao, Lin Liu, Jianguo Li, Zhenzhong Lan, and Weiyao Lin. 2025{\natexlab{a}}.
\newblock \href {https://arxiv.org/abs/2510.06133} {Creditdecoding: Accelerating parallel decoding in diffusion large language models with trace credits}.
\newblock \emph{arXiv preprint arXiv:2510.06133}.

\bibitem[{Wang et~al.(2024{\natexlab{a}})Wang, Pan, Shi, Lu, Ren, Zhou, Zhan, and Li}]{wang2024measuring}
Ke~Wang, Junting Pan, Weikang Shi, Zimu Lu, Houxing Ren, Aojun Zhou, Mingjie Zhan, and Hongsheng Li. 2024{\natexlab{a}}.
\newblock \href {https://dl.acm.org/doi/10.5555/3737916.3740930} {Measuring multimodal mathematical reasoning with math-vision dataset}.
\newblock \emph{Advances in Neural Information Processing Systems}.

\bibitem[{Wang et~al.(2025{\natexlab{b}})Wang, Fang, Jing, Shen, Shen, Wang, Ouyang, Chen, and Shen}]{wang2025time}
Wen Wang, Bozhen Fang, Chenchen Jing, Yongliang Shen, Yangyi Shen, Qiuyu Wang, Hao Ouyang, Hao Chen, and Chunhua Shen. 2025{\natexlab{b}}.
\newblock \href {https://arxiv.org/abs/2508.09138} {Time is a feature: Exploiting temporal dynamics in diffusion language models}.
\newblock \emph{arXiv preprint arXiv:2508.09138}.

\bibitem[{Wang et~al.(2025{\natexlab{c}})Wang, Xu, Jin, Jin, Zhang, and Deng}]{wang2025diffusion}
Xu~Wang, Chenkai Xu, Yijie Jin, Jiachun Jin, Hao Zhang, and Zhijie Deng. 2025{\natexlab{c}}.
\newblock \href {https://arxiv.org/abs/2508.09192} {Diffusion llms can do faster-than-ar inference via discrete diffusion forcing}.
\newblock \emph{arXiv preprint arXiv:2508.09192}.

\bibitem[{Wang et~al.(2024{\natexlab{b}})Wang, Ma, Zhang, Ni, Chandra, Guo, Ren, Arulraj, He, Jiang, Li, Ku, Wang, Zhuang, Fan, Yue, and Chen}]{wang2024mmlu}
Yubo Wang, Xueguang Ma, Ge~Zhang, Yuansheng Ni, Abhranil Chandra, Shiguang Guo, Weiming Ren, Aaran Arulraj, Xuan He, Ziyan Jiang, Tianle Li, Max Ku, Kai Wang, Alex Zhuang, Rongqi Fan, Xiang Yue, and Wenhu Chen. 2024{\natexlab{b}}.
\newblock \href {https://dl.acm.org/doi/10.5555/3737916.3740934} {Mmlu-pro: A more robust and challenging multi-task language understanding benchmark}.
\newblock \emph{Advances in Neural Information Processing Systems}.

\bibitem[{Wei et~al.(2026)Wei, Zhang, Liu, Zeng, Wang, Qi, Liu, and Zhang}]{wei2025accelerating}
Qingyan Wei, Yaojie Zhang, Zhiyuan Liu, Puyu Zeng, Yuxuan Wang, Biqing Qi, Dongrui Liu, and Linfeng Zhang. 2026.
\newblock \href {https://arxiv.org/abs/2506.10848} {Accelerating diffusion large language models with slowfast sampling: The three golden principles}.
\newblock \emph{arXiv preprint arXiv:2506.10848}.

\bibitem[{Wu et~al.(2025)Wu, Zhang, Xue, Liu, Diao, Zhu, Luo, Han, and Xie}]{wu2025fast}
Chengyue Wu, Hao Zhang, Shuchen Xue, Zhijian Liu, Shizhe Diao, Ligeng Zhu, Ping Luo, Song Han, and Enze Xie. 2025.
\newblock \href {https://arxiv.org/abs/2505.22618} {Fast-dllm: Training-free acceleration of diffusion llm by enabling kv cache and parallel decoding}.
\newblock \emph{arXiv preprint arXiv:2505.22618}.

\bibitem[{Xin et~al.(2025)Xin, Qin, Luo, Zhu, Yan, Tai, Lei, Cao, Wang, Wang, Bai, Yu, Jiang, Pu, Chen, Zhuo, He, Luo, Li, Hu, Ye, Ye, Zhang, Xu, Wang, Li, Zhai, Xue, Fu, Liu, Qiao, and Liu}]{xin2025lumina}
Yi~Xin, Qi~Qin, Siqi Luo, Kaiwen Zhu, Juncheng Yan, Yan Tai, Jiayi Lei, Yuewen Cao, Keqi Wang, Yibin Wang, Jinbin Bai, Qian Yu, Dengyang Jiang, Yuandong Pu, Haoxing Chen, Le~Zhuo, Junjun He, Gen Luo, Tianbin Li, and 13 others. 2025.
\newblock \href {https://arxiv.org/abs/2510.06308} {Lumina-dimoo: An omni diffusion large language model for multi-modal generation and understanding}.
\newblock \emph{arXiv preprint arXiv:2510.06308}.

\bibitem[{Yang et~al.(2025)Yang, Tian, Li, Zhang, Shen, Tong, and Wang}]{yang2025mmada}
Ling Yang, Ye~Tian, Bowen Li, Xinchen Zhang, Ke~Shen, Yunhai Tong, and Mengdi Wang. 2025.
\newblock \href {https://arxiv.org/abs/2505.15809} {Mmada: Multimodal large diffusion language models}.
\newblock \emph{arXiv preprint arXiv:2505.15809}.

\bibitem[{Yang et~al.(2023)Yang, Zhang, Song, Hong, Xu, Zhao, Zhang, Cui, and Yang}]{yang2023diffusion}
Ling Yang, Zhilong Zhang, Yang Song, Shenda Hong, Runsheng Xu, Yue Zhao, Wentao Zhang, Bin Cui, and Ming-Hsuan Yang. 2023.
\newblock \href {https://doi.org/10.1145/3626235} {Diffusion models: A comprehensive survey of methods and applications}.
\newblock \emph{ACM Computing Surveys}.

\bibitem[{Ye et~al.(2025)Ye, Xie, Zheng, Gao, Wu, Jiang, Li, and Kong}]{ye2025dream}
Jiacheng Ye, Zhihui Xie, Lin Zheng, Jiahui Gao, Zirui Wu, Xin Jiang, Zhenguo Li, and Lingpeng Kong. 2025.
\newblock \href {https://arxiv.org/abs/2508.15487} {Dream 7b: Diffusion large language models}.
\newblock \emph{arXiv preprint arXiv:2508.15487}.

\bibitem[{Yu et~al.(2025{\natexlab{a}})Yu, Li, and Wang}]{yu2025discrete}
Runpeng Yu, Qi~Li, and Xinchao Wang. 2025{\natexlab{a}}.
\newblock \href {https://arxiv.org/abs/2506.13759} {Discrete diffusion in large language and multimodal models: A survey}.
\newblock \emph{arXiv preprint arXiv:2506.13759}.

\bibitem[{Yu et~al.(2025{\natexlab{b}})Yu, Ma, and Wang}]{yu2025dimple}
Runpeng Yu, Xinyin Ma, and Xinchao Wang. 2025{\natexlab{b}}.
\newblock \href {https://arxiv.org/abs/2505.16990} {Dimple: Discrete diffusion multimodal large language model with parallel decoding}.
\newblock \emph{arXiv preprint arXiv:2505.16990}.

\bibitem[{Zhang et~al.(2025)Zhang, Li, Zhang, Pu, Cahyono, Hu, Liu, Zhang, Yang, Li, and Liu}]{zhang2024lmmsevalrealitycheckevaluation}
Kaichen Zhang, Bo~Li, Peiyuan Zhang, Fanyi Pu, Joshua~Adrian Cahyono, Kairui Hu, Shuai Liu, Yuanhan Zhang, Jingkang Yang, Chunyuan Li, and Ziwei Liu. 2025.
\newblock \href {https://aclanthology.org/2025.findings-naacl.51/} {{LMM}s-eval: Reality check on the evaluation of large multimodal models}.
\newblock In \emph{Findings of the Association for Computational Linguistics: NAACL 2025}.

\bibitem[{Zhang et~al.(2023)Zhang, Wang, Huang, Tasnim, and Shi}]{zhang2023survey}
Tianyi Zhang, Zheng Wang, Jing Huang, Mohiuddin~Muhammad Tasnim, and Wei Shi. 2023.
\newblock \href {https://arxiv.org/abs/2308.13142} {A survey of diffusion based image generation models: Issues and their solutions}.
\newblock \emph{arXiv preprint arXiv:2308.13142}.

\bibitem[{Zhou et~al.(2025)Zhou, Chen, Zhao, Kang, Li, Wang, Guo, Sun, Wang, Kong, Qin, and Li}]{zhou2025diffa}
Jiaming Zhou, Hongjie Chen, Shiwan Zhao, Jian Kang, Jie Li, Enzhi Wang, Yujie Guo, Haoqin Sun, Hui Wang, Aobo Kong, Yong Qin, and Xuelong Li. 2025.
\newblock \href {https://arxiv.org/abs/2507.18452} {Diffa: Large language diffusion models can listen and understand}.
\newblock \emph{arXiv preprint arXiv:2507.18452}.

\bibitem[{Zhu et~al.(2025)Zhu, Wang, Nie, Zhang, Wu, Hu, Zhou, Chen, Lin, Wen, and Li}]{zhu2025llada1.5}
Fengqi Zhu, Rongzhen Wang, Shen Nie, Xiaolu Zhang, Chunwei Wu, Jun Hu, Jun Zhou, Jianfei Chen, Yankai Lin, Ji-Rong Wen, and Chongxuan Li. 2025.
\newblock \href {https://arxiv.org/abs/2505.19223} {Llada 1.5: Variance-reduced preference optimization for large language diffusion models}.
\newblock \emph{arXiv preprint arXiv:2505.19223}.

\end{thebibliography}

\appendix
\label{sec:appendix}

\section{Related Work}
\label{related work}
\paragraph{Diffusion Large Language Model.}
Diffusion models have achieved widespread success in image and video generation~\cite{ho2020denoising,rombach2022high,melnik2024video,zhang2023survey,yang2023diffusion}, inspiring growing interest in applying them to discrete text domains. Migrating diffusion models from continuous to discrete domains presents unique challenges, and researchers have explored several technical approaches~\cite{campbell2022continuous,lovelace2023latent,yu2025discrete}. Among these, Masked Diffusion Models (MDMs) have become the most mainstream branch~\cite{austin2021structured,sahoo2024simple,chen2023cheaper,gong2024scaling}. This paradigm models text generation as an iterative “mask–denoise” process. In the forward process, noise is introduced by gradually replacing tokens in the text with a special \(\texttt{[MASK]}\) absorption state. In the reverse process, the model learns to reconstruct the original text from a fully masked sequence~\cite{ou2024your,sahoo2024simple,shi2024simplified,nie2025large}. Inspired by this framework, researchers have successfully extended MDMs to diffusion large language models (dLLMs), with LLaDA~\cite{nie2025large} and Dream~\cite{ye2025dream} attracting wide attention in the community. Their performance is competitive with autoregressive models with similar scale~\cite{yu2025discrete}.


\paragraph{Parallel Decoding.}
\label{parrel decoding}
In dLLMs, logits for all token positions are computed during each decoding iteration. Fast-dLLM identifies the root cause of degraded generation quality in parallel decoding as the violation of token dependencies under the conditional independence assumption. To address this, Fast-dLLM and Dimple propose confidence-aware parallel decoding~\cite{wu2025fast,yu2025dimple}, which dynamically adjusts the number of tokens to unmask in each iteration based on a confidence threshold. This allows generating multiple tokens simultaneously while mitigating dependency violations and preserving generation quality, thereby achieving a trade-off between quality and inference efficiency.


\paragraph{Advanced Decoding Strategies for dLLM.}
Due to the iterative denoising nature of dLLMs, their inference latency is relatively high~\cite{nie2025large}. Furthermore, as the attention in dLLMs is bidirectional, traditional KV caching is not directly applicable~\cite{wu2025fast}. This has led to the development of various acceleration strategies, including specialized KV cache mechanisms~\cite{chen2025dpad,liu2025dllm,ma2025dkv,wu2025fast,song2025sparse,huang2025mask} and advanced decoding methods~\cite{wu2025fast,dong2025saber,wang2025creditdecoding,wei2025accelerating,kong2025accelerating,kim2025klass,li2025diffusion,lu2025adablock}. Parallel decoding strategies typically sample multiple tokens at each step, thereby reducing the total number of iterative steps without retraining and achieving faster generation. Early work such as Fast-dLLM~\cite{wu2025fast} proposed a parallel decoding scheme based on a fixed confidence threshold. Saber~\cite{dong2025saber} introduced adaptive acceleration by dynamically adjusting the parallel size and employed a backtracking re-masking mechanism to correct prediction errors, improving both speed and quality. WINO~\cite{hong2025wide} implements revocable decoding in dLLMs, enabling the refinement of suspicious tokens. However, these decoding strategies are all based on semi-autoregressive decoding, where advanced decoding typically operates only within a single block per step, without leveraging already stable tokens in future blocks.

\section{More Implementation Details}
\label{sec:appendix-exper-detils}
\paragraph{Main Experiment.}
In the language domain, we employ the lm‑eval library~\footnote{\url{https://github.com/EleutherAI/lm-evaluation-harness}} to ensure evaluation consistency, stability, and reproducibility. Table~\ref{icl} reports the in‑context learning (ICL) configuration used for each benchmark.
In the vision‑language domain, we adopt the lmms‑eval~\cite{zhang2024lmmsevalrealitycheckevaluation} library to unify the evaluation of performance across multimodal understanding and reasoning benchmarks. In the audio-language domain, we follow the experimental setup of DIFFA. The generation and block lengths are set to 128 and 16, respectively. We evaluate our method on five tasks using the VoiceBench benchmark. For AlpacaEval, CommonEval, and WildVoice, we employ GPT-4o-mini as the evaluator. We set $max\_tokens$ to 1024, $temperature$ to 0.5, and $top\_p$ to 0.95.

Additionally, we present detailed inference efficiency metrics in our experiments. Among these metrics, TPF and Step focus exclusively on the decoding process of the dLLM itself. They are unaffected by hardware quantity, hardware type, GPU communication protocols, or inference frameworks, and thus directly reflect intrinsic improvements to dLLMs. All experiments are conducted on eight NVIDIA H20 GPUs.

We provide the detailed hyperparameter settings for each benchmark in Table~\ref{tab:Hyperparameters-setting}. In general, we set the historical‑consistency threshold \(\varepsilon\)=0.01 and the historical buffer length \( \mathcal{H} \) =6. Although the optimal hyperparameters vary slightly across benchmarks, their performance remains relatively consistent.

In the main experiments on LLaDA-8B-Instruct and LLaDA-1.5, we compared not only with naive decoding but also with several recent advanced decoding strategies, including PC-sample~\cite{huang2025pc}, Fast-dLLM~\cite{wu2025fast}, KLASS~\cite{kim2025klass}, and Saber~\cite{dong2025saber}. We kept their parameters consistent with the original papers to ensure fair comparison. 



\begin{table}[h]
\centering
\begin{tabular}{l c}
    \hline
    Task       & ICL \\ \hline
    HumanEval  & 0-shot   \\
    MBPP       & 3-shot   \\
    BBH        & 3-shot   \\
    MMLU-Pro   & 0-shot   \\
    TruthfulQA & 0-shot   \\
    Math       & 3-shot   \\
    Asdiv      & 0-shot   \\ \hline
    \end{tabular}
\caption{In-context learning settings for different tasks.}
\label{icl}
\end{table}

\begin{table}[t]
    \centering

    \resizebox{\columnwidth}{!}{\begin{tabular}{lcccc}
        \toprule
        \multirow{2}{*}{\textbf{Task}}
        & \multicolumn{2}{c}{\textbf{LLaDA-8B-Instruct}} 
        & \multicolumn{2}{c}{\textbf{LLaDA-1.5}} \\
        \cmidrule(lr){2-3} \cmidrule(lr){4-5}
        & $\epsilon$ & $\mathcal{H}$ & $\epsilon$ & $\mathcal{H}$ \\
        \midrule
        HumanEval   & 0.02 & 6 & 0.01  & 6 \\
        MBPP        & 0.01  & 6 & 0.01  & 4 \\
        BBH         & 0.05  & 6 & 0.01  & 6 \\
        MMLU-Pro    & 0.01  & 6 & 0.01  & 6 \\
        TruthfulQA  & 0.01  & 6 & 0.01  & 6 \\
        Math        & 0.01  & 6 & 0.001 & 7 \\
        Asdiv       & 0.01  & 6 & 0.01  & 6 \\
        \bottomrule
    \end{tabular}
    }
    \caption{Hyperparameters $\epsilon$ and $\mathcal{H}$ for different tasks.}
    \label{tab:Hyperparameters-setting}
\end{table}

\paragraph{Ablation Experiment.}
In all ablation experiments, we use LLaDA‑8B‑Instruct as the base model.  
For the ablations on generation length and block length, the hyperparameters of historical consistency threshold \(\varepsilon\) and historical buffer length \(\mathcal{H}\) are set to 0.01 and 6, respectively.  
For the ablations on \(\varepsilon\) and \(\mathcal{H}\), we fix the generation length at 256 and the block length at 32.  


\section{More Experimental Results}
\label{appendix_res}
\subsection{Scalability.}

\begin{table*}[t]
    \centering

    \begin{tabular}{l c c c c c c}
        \toprule
        \multirow{2}{*}{\textbf{Method}} &
        \multicolumn{3}{c}{\textbf{1024}} &
        \multicolumn{3}{c}{\textbf{2048}} \\
        \cmidrule(lr){2-4}\cmidrule(lr){5-7}
        &
        \emph{Score}$\uparrow$ & \emph{TPF}$\uparrow$ & \emph{Step}$\downarrow$ &
        \emph{Score}$\uparrow$ & \emph{TPF}$\uparrow$ & \emph{Step}$\downarrow$ \\
        \midrule

        {\color{vanillacol}LLaDA} &
        43.90 & 1.00 & 1024 &
        44.51 & 1.00 & 2048 \\

        PC-sampler &
        42.07 & 1.00 & 1024 &
        41.46 & 1.00 & 2048 \\

        Fast-dLLM &
        42.68 & 5.54 & 184.79 &
        44.51 & 10.50 & 195.13 \\

        KLASS &
        43.90 & 3.30 & 309.97 &
        44.51 & 6.39  & 320.63 \\

        Saber &
        43.90 & 4.87 & 210.43 &
        48.17 & 9.34 & 219.25 \\

        \rowcolor{oursbg2}\textbf{Ours} &
        \textbf{48.78} & \textbf{6.99} & \textbf{146.48} &
        \textbf{49.39} & \textbf{11.92} & \textbf{171.84} \\
        \bottomrule
    \end{tabular}
    
    \caption{Long-sequence results of LLaDA-8B-Instruct on HumanEval benchmark.}
    \label{tab:genlen_long_humaneval}
\end{table*}

Previous studies have mostly neglected the case of long‑sequence generation, typically limiting generation length to 256 or 512. To verify whether our method remains effective for long sequences, we conducted experiments with extended generation lengths. 
Specifically, the block length was set to 64 for a generation length of 1024 and 128 for a generation length of 2048, while all other parameters remained consistent with those of the main experiments. As shown in Table~\ref{tab:genlen_long_humaneval}, on the HumanEval benchmark, our method continues to be effective for generation lengths of 1024 and 2048, maintaining a stable and consistent improvement trend.


\subsection{Further Results on Additional Benchmarks.}
In this section, we present the results of LLaDA-8B-Instruct on additional benchmarks. As shown in Table~\ref{tab:gsm8k-cot}, AHD maintains superior performance and inference efficiency on GSM8K-COT~\cite{cobbe2021training} compared with naive decoding and other advanced decoding strategies, demonstrating a consistent improvement trend with the main experiments described in the text.

\begin{table*}[t]
    \centering

    \begin{tabular}{lccc}
        \toprule
        \rowcolor{headergray}
        \textbf{Method} & \emph{Score} ↑ & \emph{TPF} ↑ & \emph{Step} ↓ \\
        \midrule
        \color{vanillacol}LLaDA
        & {79.53} 
        & {1.00} 
        & {256} \\

        PC-sampler 
        & 79.98 {\scriptsize\textcolor{inc}{(+0.45)}} 
        & 1.00  
        & 256  \\
        
        Fast\_dLLM 
        & 80.29 {\scriptsize\textcolor{inc}{(+0.76)}} 
        & 3.41 {\scriptsize\textcolor{inc}{($\times$3.41)}} 
        & 75.10 {\scriptsize\textcolor{inc}{($-71\%$)}} \\
        
        KLASS 
        & 80.06 {\scriptsize\textcolor{inc}{(+0.53)}} 
        & 1.99 {\scriptsize\textcolor{inc}{($\times$1.99)}} 
        & 128.64 {\scriptsize\textcolor{inc}{($-50\%$)}} \\
        
        Saber 
        & 79.61 {\scriptsize\textcolor{inc}{(+0.08)}} 
        & 3.35 {\scriptsize\textcolor{inc}{($\times$3.35)}} 
        & 76.49 {\scriptsize\textcolor{inc}{($-70\%$)}} \\

        \cellcolor{oursbg}\textbf{Ours} 
        & \cellcolor{oursbg}\textbf{80.51} {\scriptsize\textcolor{inc}{(+0.98)}} 
        & \cellcolor{oursbg}\textbf{3.41} {\scriptsize\textcolor{inc}{($\times$3.41)}} 
        & \cellcolor{oursbg}75.06 {\scriptsize\textcolor{inc}{($-71\%$)}} \\
        \bottomrule
    \end{tabular}%
    \caption{Results of LLaDA‑8B‑Instruct on the GSM8K‑CoT benchmark.}
    \label{tab:gsm8k-cot}
\end{table*}



\subsection{Further Results from the Ablation Study.}
In this section, we provide detailed ablation studies on generation length and block length. As shown in Tables~\ref{tab:appen_gen} and ~\ref{tab:appendix_block}, we include comparisons with other advanced decoding strategies on the MMLU-Pro and BBH general task benchmarks. The comprehensive results demonstrate that AHD is robust to both generation length and block length.
Notably, as the generation length increases, the reduction in decoding steps for AHD compared to vanilla decoding becomes increasingly pronounced. 
This phenomenon can be attributed to the delayed decoding of stable tokens caused by block boundaries, which becomes more significant with longer generation lengths.


\begin{table*}[ht]
    \centering

    \resizebox{1\textwidth}{!}{\begin{tabular}{l l c c c c c c}
        \toprule
        \multirow{2}{*}{\textbf{Task}} & \multirow{2}{*}{\textbf{Method}} &
        \multicolumn{2}{c}{\textbf{128}} &
        \multicolumn{2}{c}{\textbf{256}} &
        \multicolumn{2}{c}{\textbf{512}} \\
        \cmidrule(lr){3-4}\cmidrule(lr){5-6}\cmidrule(lr){7-8}
        & &
        \emph{Score}$\uparrow$ & \emph{Step}$\downarrow$ &
        \emph{Score}$\uparrow$ & \emph{Step}$\downarrow$ &
        \emph{Score}$\uparrow$ & \emph{Step}$\downarrow$ \\
        \midrule

        \multirow{6}{*}{\textbf{MMLU-Pro}} &
        {\color{vanillacol}LLaDA} &
        {\color{vanillacol}37.42} & {\color{vanillacol}128} &
        {\color{vanillacol}35.57} & {\color{vanillacol}256} &
        {\color{vanillacol}37.46} & {\color{vanillacol}512} \\
        
        & PC-sampler &
        35.59 {\scriptsize\textcolor{dec}{(-1.83)}} & 128  &
        33.90 {\scriptsize\textcolor{dec}{(-1.67)}} & 256  &
        35.31 {\scriptsize\textcolor{dec}{(-2.15)}} & 512  \\
        
        & Fast\_dllm &
        37.36 {\scriptsize\textcolor{dec}{(-0.06)}} & 67.72 {\scriptsize\textcolor{inc}{($-47\%$)}} &
        35.70 {\scriptsize\textcolor{inc}{(+0.13)}} & 134.76 {\scriptsize\textcolor{inc}{($-47\%$)}} &
        37.32 {\scriptsize\textcolor{dec}{(-0.14)}} & 199.31 {\scriptsize\textcolor{inc}{($-61\%$)}} \\
        
        & KLASS &
        37.43 {\scriptsize\textcolor{inc}{(+0.01)}} & 88.96 {\scriptsize\textcolor{inc}{($-31\%$)}} &
        35.67 {\scriptsize\textcolor{inc}{(+0.10)}} & 187.17 {\scriptsize\textcolor{inc}{($-27\%$)}} &
        37.43 {\scriptsize\textcolor{dec}{(-0.03)}} & 293.96 {\scriptsize\textcolor{inc}{($-43\%$)}} \\
        
        & Saber &
        36.48 {\scriptsize\textcolor{dec}{(-0.94)}} & \textbf{55.65} {\scriptsize\textcolor{inc}{($-57\%$)}} &
        36.10 {\scriptsize\textcolor{inc}{(+0.53)}} & \textbf{128.83} {\scriptsize\textcolor{inc}{($-50\%$)}} &
        37.16 {\scriptsize\textcolor{dec}{(-0.30)}} & \textbf{195.01} {\scriptsize\textcolor{inc}{($-62\%$)}} \\
        
        & \cellcolor{oursbg2}\textbf{Ours} &
        \cellcolor{oursbg2}\textbf{38.12} {\scriptsize\textcolor{inc}{(+0.70)}} & \cellcolor{oursbg2}65.51 {\scriptsize\textcolor{inc}{($-49\%$)}} &
        \cellcolor{oursbg2}\textbf{37.42} {\scriptsize\textcolor{inc}{(+1.85)}} & \cellcolor{oursbg2}133.06 {\scriptsize\textcolor{inc}{($-48\%$)}} &
        \cellcolor{oursbg2}\textbf{39.00} {\scriptsize\textcolor{inc}{(+1.54)}} & \cellcolor{oursbg2}197.20 {\scriptsize\textcolor{inc}{($-61\%$)}} \\
        \midrule

        \multirow{6}{*}{\textbf{BBH}} &
        {\color{vanillacol}LLaDA} &
        {\color{vanillacol}45.71} & {\color{vanillacol}128} &
        {\color{vanillacol}53.11} & {\color{vanillacol}256} &
        {\color{vanillacol}53.14} & {\color{vanillacol}512} \\
        
        & PC-sampler &
        42.05 {\scriptsize\textcolor{dec}{(-3.66)}} & 128  &
        48.55 {\scriptsize\textcolor{dec}{(-4.56)}} & 256  &
        49.62 {\scriptsize\textcolor{dec}{(-3.52)}} & 512  \\
        
        & Fast\_dllm &
        45.37 {\scriptsize\textcolor{dec}{(-0.34)}} & 33.13 {\scriptsize\textcolor{inc}{($-74\%$)}} &
        53.17 {\scriptsize\textcolor{inc}{(+0.06)}} & 55.85 {\scriptsize\textcolor{inc}{($-78\%$)}} &
        52.81 {\scriptsize\textcolor{dec}{(-0.33)}} & 68.66 {\scriptsize\textcolor{inc}{($-87\%$)}} \\
        
        & KLASS &
        45.69 {\scriptsize\textcolor{dec}{(-0.02)}} & 59.90 {\scriptsize\textcolor{inc}{($-53\%$)}} &
        53.03 {\scriptsize\textcolor{dec}{(-0.08)}} & 98.04 {\scriptsize\textcolor{inc}{($-62\%$)}} &
        53.17 {\scriptsize\textcolor{inc}{(+0.03)}} & 125.37 {\scriptsize\textcolor{inc}{($-76\%$)}} \\
        
        & Saber &
        45.62 {\scriptsize\textcolor{dec}{(-0.09)}} & 47.10 {\scriptsize\textcolor{inc}{($-63\%$)}} &
        52.88 {\scriptsize\textcolor{dec}{(-0.23)}} & 87.31 {\scriptsize\textcolor{inc}{($-66\%$)}} &
        51.84 {\scriptsize\textcolor{dec}{(-1.30)}} & 120.27 {\scriptsize\textcolor{inc}{($-77\%$)}} \\
        
        & \cellcolor{oursbg2}\textbf{Ours} &
        \cellcolor{oursbg2}\textbf{50.07} {\scriptsize\textcolor{inc}{(+4.36)}} & \cellcolor{oursbg2}\textbf{33.11} {\scriptsize\textcolor{inc}{($-74\%$)}} &
        \cellcolor{oursbg2}\textbf{56.78} {\scriptsize\textcolor{inc}{(+3.67)}} & \cellcolor{oursbg2}\textbf{51.48} {\scriptsize\textcolor{inc}{($-80\%$)}} &
        \cellcolor{oursbg2}\textbf{59.05} {\scriptsize\textcolor{inc}{(+5.91)}} & \cellcolor{oursbg2}\textbf{60.68} {\scriptsize\textcolor{inc}{($-88\%$)}} \\
        \bottomrule
    \end{tabular}
    }
    \caption{Performance and steps on the MMLU-Pro and BBH benchmarks at various generation lengths.}
    \label{tab:appen_gen}
\end{table*}

\begin{table*}[ht]
    \centering

    \begin{tabular}{l l c c c c}
        \toprule
        \textbf{Block} & \multirow{2}{*}{\textbf{Method}} &
        \multicolumn{2}{c}{\textbf{MMLU-Pro}} &
        \multicolumn{2}{c}{\textbf{BBH}} \\
        \cmidrule(lr){3-4}\cmidrule(lr){5-6}
        \textbf{length} & &
        \emph{Score}$\uparrow$ & \emph{Step}$\downarrow$ &
        \emph{Score}$\uparrow$ & \emph{Step}$\downarrow$ \\
        \midrule

        \multirow{6}{*}{\textbf{16}} &
        {\color{vanillacol}LLaDA} &
        {\color{vanillacol}33.74} & {\color{vanillacol}256} &
        {\color{vanillacol}52.51} & {\color{vanillacol}256} \\
        
        & PC-sampler &
        33.98 {\scriptsize\textcolor{inc}{(+0.24)}} & 256  &
        48.44 {\scriptsize\textcolor{dec}{(-4.07)}} & 256 \\
        
        & Fast\_dllm &
        33.65 {\scriptsize\textcolor{dec}{(-0.09)}} & 138.76 {\scriptsize\textcolor{inc}{($-46\%$)}} &
        52.47 {\scriptsize\textcolor{dec}{(-0.04)}} & 63.37 {\scriptsize\textcolor{inc}{($-75\%$)}} \\
        
        & KLASS &
        33.73 {\scriptsize\textcolor{dec}{(-0.01)}} & 192.61 {\scriptsize\textcolor{inc}{($-25\%$)}} &
        52.57 {\scriptsize\textcolor{inc}{(+0.06)}} & 112.90 {\scriptsize\textcolor{inc}{($-56\%$)}} \\
        
        & Saber &
        34.55 {\scriptsize\textcolor{inc}{(+0.81)}} & \textbf{125.00} {\scriptsize\textcolor{inc}{($-51\%$)}} &
        51.64 {\scriptsize\textcolor{dec}{(-0.87)}} & 91.47 {\scriptsize\textcolor{inc}{($-64\%$)}} \\
        
        & \cellcolor{oursbg2}\textbf{Ours} &
        \cellcolor{oursbg2}\textbf{37.36} {\scriptsize\textcolor{inc}{(+3.62)}} & \cellcolor{oursbg2}137.02 {\scriptsize\textcolor{inc}{($-46\%$)}} &
        \cellcolor{oursbg2}\textbf{56.12} {\scriptsize\textcolor{inc}{(+3.61)}} & \cellcolor{oursbg2}\textbf{59.88} {\scriptsize\textcolor{inc}{($-77\%$)}} \\
        \midrule

        \multirow{6}{*}{\textbf{32}} &
        {\color{vanillacol}LLaDA} &
        {\color{vanillacol}35.57} & {\color{vanillacol}256} &
        {\color{vanillacol}53.11} & {\color{vanillacol}256} \\
        
        & PC-sampler &
        33.90 {\scriptsize\textcolor{dec}{(-1.67)}} & 256  &
        48.55 {\scriptsize\textcolor{dec}{(-4.56)}} & 256  \\
        
        & Fast\_dllm &
        35.70 {\scriptsize\textcolor{inc}{(+0.13)}} & 134.76 {\scriptsize\textcolor{inc}{($-47\%$)}} &
        53.17 {\scriptsize\textcolor{inc}{(+0.06)}} & 55.85 {\scriptsize\textcolor{inc}{($-78\%$)}} \\
        
        & KLASS &
        35.67 {\scriptsize\textcolor{inc}{(+0.10)}} & 187.17 {\scriptsize\textcolor{inc}{($-27\%$)}} &
        53.03 {\scriptsize\textcolor{dec}{(-0.08)}} & 98.04 {\scriptsize\textcolor{inc}{($-62\%$)}} \\
        
        & Saber &
        36.10 {\scriptsize\textcolor{inc}{(+0.53)}} & \textbf{128.83} {\scriptsize\textcolor{inc}{($-50\%$)}} &
        52.88 {\scriptsize\textcolor{dec}{(-0.23)}} & 87.31 {\scriptsize\textcolor{inc}{($-66\%$)}} \\
        
        & \cellcolor{oursbg2}\textbf{Ours} &
        \cellcolor{oursbg2}\textbf{37.42} {\scriptsize\textcolor{inc}{(+1.85)}} & \cellcolor{oursbg2}133.06 {\scriptsize\textcolor{inc}{($-48\%$)}} &
        \cellcolor{oursbg2}\textbf{56.78} {\scriptsize\textcolor{inc}{(+3.67)}} & \cellcolor{oursbg2}\textbf{51.48} {\scriptsize\textcolor{inc}{($-80\%$)}} \\
        \midrule

        \multirow{6}{*}{\textbf{64}} &
        {\color{vanillacol}LLaDA} &
        {\color{vanillacol}36.67} & {\color{vanillacol}256} &
        {\color{vanillacol}54.32} & {\color{vanillacol}256} \\
        
        & PC-sampler &
        34.13 {\scriptsize\textcolor{dec}{(-2.54)}} & 256  &
        48.49 {\scriptsize\textcolor{dec}{(-5.83)}} & 256  \\
        
        & Fast\_dllm &
        36.79 {\scriptsize\textcolor{inc}{(+0.12)}} & 133.33 {\scriptsize\textcolor{inc}{($-48\%$)}} &
        53.40 {\scriptsize\textcolor{dec}{(-0.92)}} & 51.05 {\scriptsize\textcolor{inc}{($-80\%$)}} \\
        
        & KLASS &
        36.82 {\scriptsize\textcolor{inc}{(+0.15)}} & 185.64 {\scriptsize\textcolor{inc}{($-27\%$)}} &
        53.54 {\scriptsize\textcolor{dec}{(-0.78)}} & 89.47 {\scriptsize\textcolor{inc}{($-65\%$)}} \\
        
        & Saber &
        37.28 {\scriptsize\textcolor{inc}{(+0.61)}} & \textbf{124.10} {\scriptsize\textcolor{inc}{($-52\%$)}} &
        54.32 {\scriptsize\textcolor{dec}{(-0.0)}} & 86.45 {\scriptsize\textcolor{inc}{($-66\%$)}} \\
        
        & \cellcolor{oursbg2}\textbf{Ours} &
        \cellcolor{oursbg2}\textbf{38.02} {\scriptsize\textcolor{inc}{(+1.35)}} & \cellcolor{oursbg2}131.54 {\scriptsize\textcolor{inc}{($-49\%$)}} &
        \cellcolor{oursbg2}\textbf{55.57} {\scriptsize\textcolor{inc}{(+1.25)}} & \cellcolor{oursbg2}\textbf{48.93} {\scriptsize\textcolor{inc}{($-81\%$)}} \\
        \midrule

        \multirow{6}{*}{\textbf{128}} &
        {\color{vanillacol}LLaDA} &
        {\color{vanillacol}36.67} & {\color{vanillacol}256} &
        {\color{vanillacol}53.46} & {\color{vanillacol}256} \\
        
        & PC-sampler &
        34.06 {\scriptsize\textcolor{dec}{(-2.61)}} & 256  &
        48.55 {\scriptsize\textcolor{dec}{(-4.91)}} & 256 \\
        
        & Fast\_dllm &
        36.74 {\scriptsize\textcolor{inc}{(+0.07)}} & 134.67 {\scriptsize\textcolor{inc}{($-47\%$)}} &
        52.48 {\scriptsize\textcolor{dec}{(-0.98)}} & 48.92 {\scriptsize\textcolor{inc}{($-81\%$)}} \\
        
        & KLASS &
        36.71 {\scriptsize\textcolor{inc}{(+0.04)}} & 187.96 {\scriptsize\textcolor{inc}{($-27\%$)}} &
        52.66 {\scriptsize\textcolor{dec}{(-0.80)}} & 85.47 {\scriptsize\textcolor{inc}{($-67\%$)}} \\
        
        & Saber &
        37.14 {\scriptsize\textcolor{inc}{(+0.47)}} & \textbf{126.63} {\scriptsize\textcolor{inc}{($-51\%$)}} &
        53.11 {\scriptsize\textcolor{dec}{(-0.35)}} & 89.07 {\scriptsize\textcolor{inc}{($-65\%$)}} \\
        
        & \cellcolor{oursbg2}\textbf{Ours} &
        \cellcolor{oursbg2}\textbf{37.84} {\scriptsize\textcolor{inc}{(+1.17)}} & \cellcolor{oursbg2}132.14 {\scriptsize\textcolor{inc}{($-48\%$)}} &
        \cellcolor{oursbg2}\textbf{53.49} {\scriptsize\textcolor{inc}{(+0.03)}} & \cellcolor{oursbg2}\textbf{48.08} {\scriptsize\textcolor{inc}{($-81\%$)}} \\
        \bottomrule
    \end{tabular}
    \caption{Performance and inference speed on the MMLU-Pro and BBH benchmarks at different block lengths.}
    \label{tab:appendix_block} 
\end{table*}

\section{Comparison with Other Works}
In this section, we focus on the differences between AHD and other decoding methods. Although KLASS~\cite{kim2025klass} also introduces the concept of KL divergence, it is important to emphasize that AHD is fundamentally different from the step-wise KL adopted in KLASS, which only compares adjacent distributions (i.e., $p^{t}$ and $p^{t+1}$). In contrast, AHD fixes the current distribution $P_j^t$ as an anchor and retrospectively compares it against the predictions along the entire historical trajectory.

On the one hand, such an anchor-based tracking mechanism provides a more direct characterization of whether the current prediction remains consistently stable over a period of time, rather than merely reflecting local fluctuations between two neighboring steps, thereby avoiding misleading judgments caused by local fluctuations. On the other hand, by performing anchored real-time backtracking, AHD can more sensitively capture the emergence of the absolute stability trend and thus avoid unnecessary iterative decoding steps.

Moreover, KLASS is mainly applied to decoding within the current block, whereas AHD is designed for decoding in future blocks. Therefore, our method is substantially different from the KLASS strategy that computes KL divergence only between adjacent decoding steps.


Although Prophet~\cite{li2025diffusion} also mentions the phenomenon of delayed decoding, the phenomenon it describes is fundamentally different from the one discussed in our paper. Our focus is on the delayed decoding of cross-block tokens caused by block boundaries under the semi-autoregressive decoding framework. In contrast, Prophet considers the situation where certain tokens can already be confidently determined at an early stage of decoding, and then uses dynamic monitoring of confidence gaps to terminate the iterative decoding process early. Therefore, both the phenomenon we investigate and the motivation of our method are fundamentally different from those of Prophet.

\section{Theoretical Analysis}
\label{Theoretical}

\subsection{Assumption and Notation}

\paragraph{Token-level Prediction Trajectory.}
Let $\mathcal{V}$ be the vocabulary with $|\mathcal{V}|$ tokens, and $\Delta^{|\mathcal{V}|-1}$ denote the probability simplex. For a fixed position $j$ (we sometimes omit $j$ when clear from context), dLLM produces a token-level prediction distribution at reverse step $t=0,\dots,T$:
\begin{equation}
    P^{t}_{j} \in \Delta^{|\mathcal{V}|-1}.
\end{equation}

\paragraph{Historical buffer and exponential weights.}
In AHD, we maintain a \emph{historical buffer} of length $H$ containing recent predictions for each position $j$:
\begin{equation}
    \mathcal{H}_j^t = \{P^{t-H+1}_{j},\dots,P^{t-1}_{j},P^t_{j}\},
\end{equation}
where we only keep the most recent $H$ steps and assume $t \ge H$ in practice.

For a given $\lambda>0$, time weights are defined using an exponential decay function as follows:
\begin{equation}
    w_\tau = \frac{\exp(-\lambda \tau)}{\sum_{k=1}^{H-1} \exp(-\lambda k)},\quad \tau = 1,\dots,H-1,
\end{equation}
where $w_\tau>0$ and $\sum_{\tau=1}^{H-1} w_\tau = 1$. Moreover, due to exponential decay, we have:
\begin{equation}
    w_1 = \max_{\tau \in \{1,\dots,H-1\}} w_\tau.
\end{equation}
This property will be used to relate the trajectory-level consistency score $D^t_{j}{(\mathrm{acs})}$ to the one-step KL divergence.

\paragraph{Token Embeddings and Semantic Representations.}
Let $E(v)\in\mathbb{R}^d$ denote the embedding of token $v\in\mathcal{V}$.
\textbf{Assumption 1 (Bounded embeddings).}
There exists $R>0$ such that for all $v\in\mathcal{V}$:
\begin{equation}
    \|E(v)\|_2 \le R.
\end{equation}
In practice, the predictive distribution at each position is computed from high-dimensional hidden states in a Transformer network. For theoretical analysis, we abstract these hidden states by the following \emph{semantic representation}:
\begin{equation}
    \mathbf{e}_j^t = \sum_{v\in\mathcal{V}} P_j^{t}(v)\,E(v) \in \mathbb{R}^d.
    \label{et}
\end{equation}
This representation can be viewed as a proxy for the position-wise hidden state in a lower-dimensional embedding space. Our results rely only on the boundedness of such representations (Assumption~1), not on the exact form of the network.

\paragraph{One-step KL and Anchor-based Historical Consistency.}
In Section~\ref{AHD} of the main text, we define the one-step KL divergence for position $j$ at step $t$ as:
\begin{equation}
\begin{split}
    D^t_{j}(\mathrm{one\mbox{-}step})
    &= D_{\mathrm{KL}}\!\left(P_j^{t} \,\Vert\, P_{j}^{t-1}\right) \\
    &= \sum_{v\in\mathcal{V}} P_j^{t}(v) \log \frac{P_j^{t}(v)}{P_{j}^{t-1}(v)}.
\end{split}
\end{equation}
To extend beyond single-step behavior, AHD introduces an \emph{anchor-based historical consistency} criterion. Specifically, at step $t$, the current prediction $P_t^{j}$ for position $j$ is designated as the \emph{anchor}:
\begin{equation}
    P_{j,anchor}^{t} =  P_j^{t}.
\end{equation}
For each backtracking offset $\tau\in\{1,\dots,H-1\}$, the anchored KL divergence is:
\begin{equation}
    \delta_{j}^{t,\tau} =  D_{\mathrm{KL}}\!\left(P_{j,anchor}^{t} \,\Vert\, P_{j}^{t-\tau}\right),
\end{equation}
and the corresponding \emph{historical consistency sequence} is: 
\begin{equation}
    \{\delta_{j}^{t,\tau},\dots,\delta_{j}^{t,H-1}\}.
\end{equation}

\noindent\textbf{Definition 1 (Anchor-based historical consistency score).}
For position $j$ at step $t$, we define the \emph{anchor-based historical consistency score} as an exponentially weighted sum:
\begin{equation}
\begin{split}
    D_{j}^{t}(acs)
    &=
    \sum_{\tau=1}^{H-1} w_\tau \,\delta_{j}^{t,\tau} \\
    &=
    \sum_{\tau=1}^{H-1} w_\tau \,
    D_{\mathrm{KL}}\!\left(P_{j,anchor}^{t} \,\Vert\, P_{j}^{t-\tau}\right).
\end{split}
\label{acs-score}
\end{equation}
Intuitively, $D_{j}^{t}(acs)$ measures how consistent the current ``anchor'' distribution $P_t^{j}$ is with the entire recent prediction trajectory stored in the historical buffer, rather than with a single previous step. 

\subsection{Anchor-based Historical Consistency Controls Distributional and Semantic Stability}
We first demonstrate that when the anchor-based historical consistency score $D_{j}^{t}(acs)$ is small, the current prediction $P_j^{t}$ and its semantic representation $\mathbf e_j^{t}$ are both close to the exponentially weighted historical mean. This justifies $D_{j}^{t}(acs)$ as a \emph{trajectory-level} stability metric.
For notational simplicity, we drop the position index $j$ in this subsection.

\paragraph{Historical Mean Distribution and Embedding.}
First of all, we define the historical exponential average distribution as:
\begin{equation}
    \bar{P}_t = \sum_{\tau=1}^{H-1} w_\tau P_{t-\tau},
\end{equation}
and the corresponding semantic average as:
\begin{equation}
    \bar{\mathbf{e}}_t = \sum_{\tau=1}^{H-1} w_\tau \mathbf{e}_{t-\tau}.
    \label{average et}
\end{equation}

\paragraph{\textbf{Lemma 1 (Anchor-based historical consistency controls deviation from the historical mean distribution).}}
For $t\ge H$, we have:
\begin{equation}
    \|P_t - \bar{P}_t\|_1 \le \sqrt{2 D_{t}{(acs)}},
\label{Lemma1}
\end{equation}
where $D_t(acs)$ is defined as in ~Equation~\eqref{acs-score}.

\noindent\textbf{\emph{Proof.}}
We use the standard Pinsker inequality, which states that for any distributions $P,Q$:
\begin{equation}
    \|P - Q\|_1 \le \sqrt{2 D_{\mathrm{KL}}(P\Vert Q)}.
\end{equation}
Applying this with $P=P_t$ and $Q=P_{t-\tau}$, we obtain:
\begin{equation}
\begin{split}
    \|P_t - P_{t-\tau}\|_1
    \le
    \sqrt{2 D_{\mathrm{KL}}(P_t\Vert P_{t-\tau})},\\
    \quad \tau = 1,\dots,H-1.
\end{split}
\label{condition}
\end{equation}
On the other hand, we have:
\begin{equation}
\begin{aligned}
    \|P_t - \bar{P}_t\|_1
    &= \Big\| P_t - \sum_{\tau=1}^{H-1} w_\tau P_{t-\tau} \Big\|_1 \\
    &= \Big\| \sum_{\tau=1}^{H-1} w_\tau (P_t - P_{t-\tau}) \Big\|_1 \\
    &\le \sum_{\tau=1}^{H-1} w_\tau \|P_t - P_{t-\tau}\|_1 \\
    &\quad\text{(triangle inequality)}\\
    &\le \sum_{\tau=1}^{H-1} w_\tau
    \sqrt{2 D_{\mathrm{KL}}(P_t\Vert P_{t-\tau})}\\
    &\quad\text{(by equation\eqref{condition})}.
\end{aligned}
\label{triangle}
\end{equation}
Since $f(x)=\sqrt{x}$ is concave on $\mathbb{R}_+$, using Jensen's inequality yields:
\begin{equation}
\begin{aligned}
    &\sum_{\tau=1}^{H-1} w_\tau 
    \sqrt{2 D_{\mathrm{KL}}(P_t\Vert P_{t-\tau})} 
    \\
    &\le \sqrt{2 \sum_{\tau=1}^{H-1} w_\tau D_{\mathrm{KL}}(P_t\Vert P_{t-\tau})} \\
    &=\sqrt{2 D_{t}{(acs)}}.
\end{aligned}
\label{Jensen}
\end{equation}
Combining Equations~\eqref{triangle} and~\eqref{Jensen} gives Equation~\eqref{Lemma1}. \hfill$\square$

\paragraph{\textbf{Lemma 2 (Anchor-based historical consistency controls deviation in embedding space).}}
Under Assumption 1, for $t\ge H$, we have:
\begin{equation}
    \|\mathbf{e}_t - \bar{\mathbf{e}}_t\|_2
    \le
    R \sqrt{2 D_{t}{(acs)}}.
    \label{Lemma 2}
\end{equation}

\emph{Proof.}
By the definitions \eqref{et} and \eqref{average et},
\begin{equation}
    \mathbf{e}_t = \sum_v P_t(v)E(v),\quad
    \bar{\mathbf{e}}_t = \sum_v \bar{P}_t(v)E(v).
\end{equation}
Hence,
\begin{equation}
    \mathbf{e}_t - \bar{\mathbf{e}}_t
    = \sum_v (P_t(v) - \bar{P}_t(v)) E(v).
\end{equation}
Taking the $\ell_2$ norm and using the triangle inequality and Assumption 1,
\begin{equation}
\begin{aligned}
    \|\mathbf{e}_t - \bar{\mathbf{e}}_t\|_2
    &\le \sum_v |P_t(v) - \bar{P}_t(v)| \cdot \|E(v)\|_2 \\
    &\le R \sum_v |P_t(v) - \bar{P}_t(v)| \\
    &= R \|P_t - \bar{P}_t\|_1.
\end{aligned}
\end{equation}
By \textbf{Lemma 1}, $\|P_t - \bar{P}_t\|_1 \le \sqrt{2 D_{t}{(acs)}}$, thus \eqref{Lemma 2} follows. \hfill$\square$

Combining Lemmas 1 and 2 show that small anchor-based historical consistency $D_t{(acs)}$ implies that the current prediction is close to an exponentially weighted average over its recent trajectory, both in distribution and embedding space. This is one aspect of the \emph{trajectory-level} stability captured by AHD.

\subsection{Anchor-base Historical Consistency is Sensitive to Long-range Oscillations}
We demonstrate that the anchor-based score $D_{t}{(acs)}$ is sensitive to long-range oscillations, whereas the one-step KL $D_t{(one\mbox{-}step)}$ can remain small under such behavior.

\paragraph{\textbf{Lemma 3.}}
Suppose there exists some $\tau^\star \in \{1,\dots,H-1\}$ such that:
\begin{equation}
    D_{\mathrm{KL}}(P_t \Vert P_{t-\tau^\star}) \ge \Delta > 0.
\end{equation}
Then,
\begin{equation}
    D_{t}^{(\mathrm{acs})} \ge w_{\tau^\star} \Delta.
\end{equation}

\noindent\textbf{\emph{Proof.}}
By definition \eqref{acs-score},
\begin{equation}
\begin{aligned}
    D_{t}{(acs)}
    &=
    \sum_{\tau=1}^{H-1} w_\tau D_{\mathrm{KL}}(P_t\Vert P_{t-\tau}) \\
    &\ge
    w_{\tau^\star} D_{\mathrm{KL}}(P_t\Vert P_{t-\tau^\star}) \\
    &\ge
    w_{\tau^\star}\Delta.
\end{aligned}
\end{equation}
Therefore, as long as the current distribution $P_t$ differs significantly from at least one historical point in the buffer, $D_{t}{(acs)}$ cannot be arbitrarily small. In contrast, the one-step KL $D_t{(one\mbox{-}step)} = D_{\mathrm{KL}}(P_t \Vert P_{t-1})$ may remain near zero under long-period oscillations (e.g., with period greater than 1), failing to capture instability on a longer time scale. \hfill$\square$

\section{Case Study}
In this section, we present several detailed cases. By unlocking stable tokens across blocks in advance, AHD significantly reduces the number of decoding steps while maintaining high accuracy. In contrast, the naive decoding strategy of LLaDA performs redundant inference and even produces incorrect answers.


\begin{figure*}[h]
    \centering
    \begin{tcolorbox}[
        title=\texttt{Case E.1: Asdiv, Generation Length 256, Block Length 32},
        width=\textwidth 
    ]
    \begin{flushleft}
        {\textbf{Question:}} Given the following problem, reason and give a final answer to the problem. Problem: If Tom plans to fix 158 watches at the rate of 12 watches per day, how many watches will he fix on the last day? Your response should end with "The final answer is [answer]" where [answer] is the response to the problem.  \\
        \textbf{Ground-truth answer:} \textcolor{red}{2} \\ 
        \vspace{1em}

        \rule{\textwidth}{0.4pt} 
        \textbf{LLaDA-8B-Instruct (decoding step: \textcolor{darkblue}{256})}
        To determine how many watches Tom will fix on the last day, we need to follow these steps:
        
        \begin{enumerate}
            \item Calculate the total number of days Tom will be fixing watches.
            \item Determine the total number of watches Tom will fix in the full days.
            \item Subtract the number of watches fixed in the full days from the total number of watches to find out how many watches he will fix on the last day.
        \end{enumerate}
        
        First, let's calculate the total number of days Tom will be fixing watches. Since he plans to fix 158 watches at a rate of 12 watches per day, we divide the total number of watches by the rate:
        
        \[
        \frac{158}{12} = 13
        \]
        
        This means Tom will be fixing watches for 13 days.
        
        Next, we calculate the total number of watches Tom will fix in the full days:
        
        \[
        13 \times 12 = 156
        \]
        
        Finally, we find out how many watches Tom will fix on the last day:
        
        \[
        158 - 156 = 12
        \]
        
        The final answer is \textcolor{red}{12}.

        \rule{\textwidth}{0.4pt} 
        \textbf{AHD (decoding step: \textcolor{darkblue}{100})}

        To determine how many watches Tom will fix on the last day, we need to follow these steps:

        \begin{enumerate}
            \item Calculate the total number of watches Tom plans to fix.
            \item Determine how many days it will take him to fix the watches at the rate of 12 watches per day.
            \item Find out how many watches he will fix on the last day.
        \end{enumerate}
        
        First, we know the total number of watches Tom plans to fix is 158.
        
        Next, we need to calculate how many days it will take to fix 158 watches at the rate of 12 watches per day. We do this by dividing the total number of watches by the rate:
        
        \[ 158 \div 12 = 13 \text{ R } 2 \]
        
        This means Tom will need 13 days to fix 158 watches, and he will have 2 watches left to fix on the last day.
        
        Therefore, the number of watches Tom will fix on the last day is 2.
        
        The final answer is \textcolor{red}{2}.
        
    \end{flushleft}
    \end{tcolorbox}
\end{figure*}

\begin{figure*}[t]
\begin{tcolorbox}[
        title={Case E.2: HumanEval, Generation Length 256, Block Length 32},width=\textwidth]
    \begin{flushleft}
        \textbf{Question:} 
        \scriptsize
\begin{verbatim}
def triples_sum_to_zero(l: list):
    """
    triples_sum_to_zero takes a list of integers as an input.
    it returns True if there are three distinct elements in the list that
    sum to zero, and False otherwise.

    >>> triples_sum_to_zero([1, 3, 5, 0])
    False
    >>> triples_sum_to_zero([1, 3, -2, 1])
    True
    >>> triples_sum_to_zero([1, 2, 3, 7])
    False
    >>> triples_sum_to_zero([2, 4, -5, 3, 9, 7])
    True
    >>> triples_sum_to_zero([1])
    False
    """
\end{verbatim}
        
        
        \rule{\textwidth}{0.4pt} 
        \textbf{LLaDA-8B-Instruct (decoding step: \textcolor{darkblue}{256})}
\begin{verbatim}
def triples_sum_to_zero(l: list):
    """
    triples_sum_to_zero takes a list of integers as an input.
    it returns True if there are three distinct elements in the list that
    sum to zero, and False otherwise.

    >>> triples_sum_to_zero([1, 3, 5, 0])
    False
    >>> triples_sum_to_zero([1, 3, -2, 1])
    True
    >>> triples_sum_to_zero([1, 2, 3, 7])
    False
    >>> triples_sum_to_zero([2, 4, -5, 3, 9, 7])
    True
    >>> triples_sum_to_zero([1])
    False
    """
    l.sort()
    for i in range(len(l) - 2):
        for j in range(i + 1, len(l)):
            if l[i] + l[j] + l[j + 1] == 0:
                return True
    return False

# Test cases
if __name__ == "__main__":
    import doctest

    doctest.testmod()
\end{verbatim}

        \rule{\textwidth}{0.4pt} 
        \textbf{AHD (decoding step: \textcolor{darkblue}{29})}
        \begin{verbatim}
    def triples_sum_to_zero(l: list):
    """
    triples_sum_to_zero takes a list of integers as an input.
    it returns True if there are three distinct elements in the list that
    sum to zero, and False otherwise.

    >>> triples_sum_to_zero([1, 3, 5, 0])
    False
    >>> triples_sum_to_zero([1, 3, -2, 1])
    True
    >>> triples_sum_to_zero([1, 2, 3, 7])
    False
    >>> triples_sum_to_zero([2, 4, -5, 3, 9, 7])
    True
    >>> triples_sum_to_zero([1])
    False
    """
    n = len(l)
    for i in range(n):
        for j in range(i + 1, n):
            for k in range(j + 1, n):
                if l[i] + l[j] + l[k] == 0:
                    return True
    return False

import doctest
if __name__ == "__main__":
    doctest.testmod()
        \end{verbatim}

    \end{flushleft}
\end{tcolorbox}
\end{figure*}

\begin{figure*}[h]
    \centering
    \begin{tcolorbox}[
        title=\texttt{Case E.3: MMLU-Pro, Generation Length 256, Block Length 32},
        width=\textwidth 
    ]
    \begin{flushleft}
    
    {\textbf{Question:}} 
    \footnotesize
    The following are multiple choice questions (with answers) about math. Think step by step and then finish your answer with "the answer is (X)" where X is the correct letter choice.

    Question: \\
    A group of scientists wanted to estimate the proportion of geese returning to the same site for the next breeding season. Suppose they decided to increase the sample size from 200 to 2,000. How will this affect the distribution of the sample proportion?

    Options: \\
    A. The shape of the distribution of the sample proportion will change significantly. \\
    B. There will be no noticeable change in the distribution of the sample proportion. \\
    C. The distribution of the sample proportion will become more uniform. \\
    D. The distribution of the sample proportion will become skewed. \\
    E. The distribution of the sample proportion will be more spread out. \\
    F. The distribution of the sample proportion will more closely resemble the binomial distribution. \\
    G. The distribution of the sample proportion will be less spread out. \\
    H. The spread of the distribution of the sample proportion will remain unaffected.

    Answer: Let's think step by step.

    A: Let's think step by step.
    
        \textbf{Ground-truth answer:} \textcolor{red}{G} \\ 
        \vspace{1em}

        \rule{\textwidth}{0.4pt} 
        \textbf{LLaDA-8B-Instruct (decoding step: \textcolor{darkblue}{256})}
        
    To answer this question, let's consider the effect of increasing the sample size on the distribution of the sample proportion.

    1. \textbf{Understanding the Distribution of the Sample Proportion:} \\
    - The distribution of the sample proportion is described by the sampling distribution of the binomial proportion. \\
    - As the sample size increases, the sampling distribution of the binomial proportion converges to a normal distribution. This is known as the Central Limit Theorem.

    2. \textbf{Effect of Increasing Sample Size:} \\
    - Increasing the sample size from 200 to 2,000 will make the sampling distribution of the binomial proportion more closely resemble a normal distribution. \\
    - This means that the distribution of the sample proportion will become more spread out (i.e., it will have a wider range of values).

    3. \textbf{Conclusion:} \\
    - The correct answer is E. The distribution of the sample proportion will be more spread out.

    The answer is (\textcolor{red}{E}).

        \rule{\textwidth}{0.4pt} 
        \textbf{AHD (decoding step: \textcolor{darkblue}{157})}
        
    To answer this question, let's consider the effect of increasing the sample size on the distribution of the sample proportion.

    1. \textbf{Understanding the Distribution of Sample Proportion:} \\
    - The distribution of the sample proportion is influenced by the sample size. \\
    - As the sample size increases, the distribution of the sample proportion tends to become more closely related to the population proportion.

    2. \textbf{Effect of Increasing Sample Size:} \\
    - When the sample size is increased from 200 to 2,000, the distribution of the sample proportion will become more concentrated around the population proportion. \\
    - This is because a larger sample size provides more information about the population, leading to a more accurate estimate of the population proportion.

    3. \textbf{Conclusion:} \\
    - As the sample size increases, the distribution of the sample proportion will become less spread out. \\
    - This is because the sample proportion will be closer to the population proportion, resulting in a narrower range of possible sample proportions.

    Therefore, the correct description of how the sample size affects the distribution of the sample proportion is: \\
    G. The distribution of the sample proportion will be less spread out.

    The answer is (\textcolor{red}{G}).
    \end{flushleft}
    \end{tcolorbox}
\end{figure*}

\section{LLM Usage}
This section clarifies the role of large language models (LLMs) in this study. LLMs are used only for language refinement, including grammar and readability. They do not participate in any aspect of research design.


\end{document}